\documentclass[12pt]{article}
\pdfoutput=1

\usepackage[round]{natbib}

\usepackage{amsmath,amssymb,amsfonts}
\usepackage{caption}
\usepackage{subcaption}
\usepackage{float}
\usepackage{algorithmic}
\usepackage{textcomp}
\usepackage{xcolor}
\usepackage{times}
\usepackage{graphicx}
\usepackage{color}
\usepackage{multirow}
\usepackage{rotating}
\usepackage{bbm}
\usepackage{latexsym}

\usepackage{subfiles}

\DeclareMathOperator\erf{erf}
\newcommand\numberequation{\addtocounter{equation}{1}\tag{\theequation}}

\title{ Prototype Analysis in Hopfield Networks with Hebbian Learning }
\author{ Hayden McAlister$^{1}$, Anthony Robins$^{1}$, Lech Szymanski$^{1}$ }
\date{
  $^{1}$School of Computing, University of Otago, Dunedin, New Zealand.
}

\begin{document}
\maketitle
\pagebreak


\begin{center} {\bf Abstract} \end{center}
We discuss prototype formation in the Hopfield network. Typically, Hebbian learning with highly correlated states leads to degraded memory performance. We show this type of learning can lead to prototype formation, where unlearned states emerge as representatives of large correlated subsets of states, alleviating capacity woes. This process has similarities to prototype learning in human cognition. We provide a substantial literature review of prototype learning in associative memories, covering contributions from psychology, statistical physics, and computer science.  We analyze prototype formation from a theoretical perspective and derive a stability condition for these states based on the number of examples of the prototype presented for learning, the noise in those examples, and the number of non-example states presented. The stability condition is used to construct a probability of stability for a prototype state as the factors of stability change. We also note similarities to traditional network analysis, allowing us to find a prototype capacity. We corroborate these expectations of prototype formation with experiments using a simple Hopfield network with standard Hebbian learning. We extend our experiments to a Hopfield network trained on data with multiple prototypes and find the network is capable of stabilizing multiple prototypes concurrently. We measure the basins of attraction of the multiple prototype states, finding attractor strength grows with the number of examples and the agreement of examples. We link the stability and dominance of prototype states to the energy profile of these states, particularly when comparing the profile shape to target states or other spurious states.

\section{Introduction}

Associative memories attempt to store a set of states, represented as vectors of either continuous or binary values. After learning, models are presented with a probe state, iterating to some final state hopefully among the learned set. The details of the storage/learning rule, update method, and other properties vary, but typically a model is only successful when learned states dominate the attractor space and are recalled with wide basins of attraction. Associative memories often suffer from low capacities and the appearance of spurious states, stable but unlearned states that may disrupt recall. Spurious states become dominant as the number of learned states approaches the capacity, resulting in a model that almost never recalls a learned state. We demonstrate the formation of prototype states, a category of unlearned state that represent a large subset of correlated learned states. Prototype states allow many learned states to be forgotten, easing capacity issues by collapsing many attractors into a single strong basin.

Perhaps the most studied model of associative memory is the Hopfield network \citep{Hopfield1982, Hopfield1984}. The traditional Hopfield network learns a set of states using Hebbian learning \citep{Hebb1949}, and an update rule defined by hard-limiting the product of weight matrix and state repeatedly. Hopfield showed this process will always terminate in a stable state as long as the neurons are updated asynchronously, and the weight matrix is symmetrical \citep{Hopfield1982}. We formalize the Hopfield network further in Section \ref{Section:Formalization}. The Hopfield network is not immune to the common problems of associative memory, with well-studied capacities for both orthogonal and random states \citep{Hertz1991, Bruck1988, McEliece1987}, and proofs that the number of spurious states increases exponentially with states stored \citep{Bruck1990}. The issues discussed above are particularly acute in the case of learned states with high correlation, such as examples drawn from some shared class. These states tend to result in crosstalk in the weight matrix, interfering with the basins of attraction and resulting in numerous spurious states, poor storage, and poor retrieval.

\begin{figure}[H]
    \includegraphics[width=\textwidth]{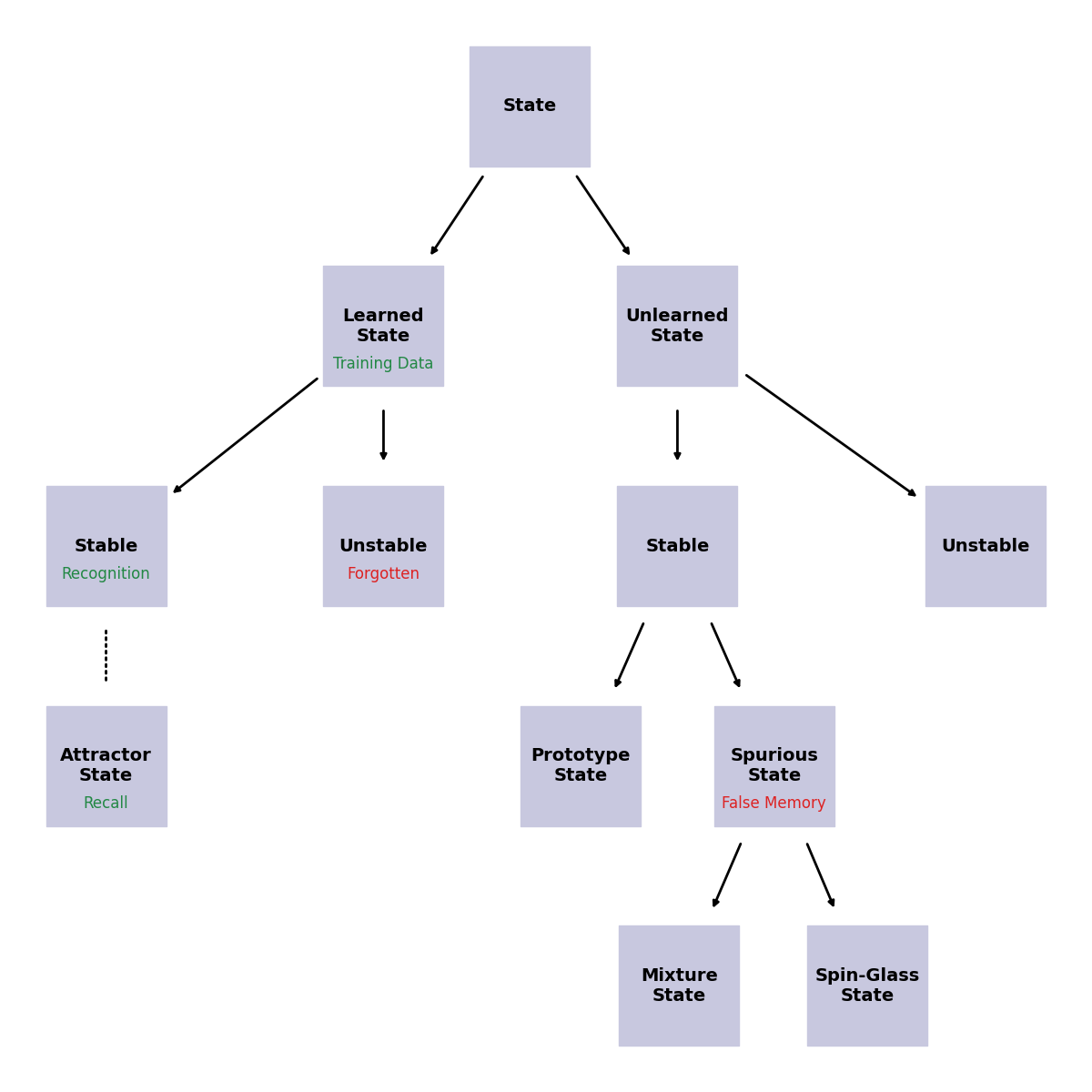}
    \caption{The taxonomy of states in an associative memory. The relationship between stable learned states and attractor states is continuous, measuring the attractor strength. Our work focuses on the distinction between the classes of stable unlearned states; separating prototypes from other spurious states.}
    \label{Fig:TaxonomyOfStates}
\end{figure}

Figure \ref{Fig:TaxonomyOfStates} shows the taxonomy of states in the context of the Hopfield network, included to standardize language in our discussion. Associative memories have been approached from many perspectives and fields, resulting in several terms for the same concepts. We will use the terminology in Figure \ref{Fig:TaxonomyOfStates} to avoid confusion: networks learn a set of learned states in the aim to stabilize them. Strongly stabilized states become attractor states, which can be recalled (nearby probes iterate to them). Weakly stabilized states may only be recognized (when probed with the state it does not iterate away, but nearby probes may not iterate to it). Any learned states not stabilized are forgotten. States that are not in the learned set but are stable may be spurious, which have been widely studied and generally fall into the category of mixture states or spin-glass states (for a distinction, see \citep{Haykin2009}).

While most works have focused on learned states, particularly on the transition from learned stable to learned unstable states, there are useful distinctions on the unlearned side as well. Typically, all unlearned stable states are regarded as actively harmful to an associative memory. However, we show that some unlearned stable states form as representatives for collections of learned states, allowing potentially many learned states to be forgotten and coalesce into a single strong attractor. In this sense, we call these states ``prototypes'', reminiscent of prototype learning in psychology \citep{Rosch1973}. Prototype states have distinct storage properties in the Hopfield network, which can be used to distinguish them from spurious states \citep{Gorman2017}. In contrast to spurious states, prototype states are useful to the associative memory, alleviating the problems of a near-capacity network. In this paper, we set out the conditions in which prototype states form, show how the Hebbian learning rule gives rise to prototype formation, and look at the transition from individual memory to prototype storage. We also present experiments that confirm the expectations of our theory, including investigations into the location of prototype states, and the dominance of prototype states in the attractor space.

\section{Prior Work}
\label{Sec:LiteratureReview}

\subsection{Psychology}

The field of psychology has discussed prototypes in the context of human cognition extensively, particularly stemming from Rosch's work \citep{Rosch1973} which looked at the ability for humans to learn prototypes without prior exposure. It was found humans tend to learn items both quicker and easier when presented with other examples from the same category. Rosch also found that humans could identify the item from a collection that was most representative of a category, indicating some sort of prototype learning occurs in human cognition as well as a measure of similarity between examples and prototypes. If the associative memories were to act similarly to human cognition we would expect it too to form prototypes, and perhaps learn examples and representatives more quickly. We will limit the scope of this paper to simply investigating prototype formation, rather than ease or speed of learning which will require further tools for analysis.

\subsection{Prototype-Based Models}

Some early models of associative memory operate on prototypes directly. Kruschke writes extensively about, and provides a good summary of, these historical models \citep{Kruschke2008}. Of particular interest, Kruschke discusses aspects of category-prototype learning in artificial neural networks. The simplest family of these prototype-specific models are the Exemplar models, in which all example items are stored persistently in a list alongside a category label. Recall is performed by comparing a probe to all examples and aggregating the similarities, with a category assigned based on the highest aggregate similarity. Exemplar models are a crude form of prototype models, with little to no update dynamics and a biologically implausible storage mechanism. Furthermore, categories must be determined beforehand, meaning states drawn from a shared class may be categorized differently. This supervised approach may lead to degradation of the prototype representations and model performance.

Another family of early prototype-specific models are the Aggregation models, where learned states are aggregated together by the learning process to form a representative of a category. Gluck and Bower implement one such Aggregation model as a single-hidden-layer feed-forward network \citep{GluckBower1988}, effectively forming prototypes as weights in the hidden layer of the network. These models are more interesting than Exemplar models as they allow nonlinear network dynamics to aid in updates rather than simple similarity scores, although biological plausibility is still lacking. Simple prototype models usually classify items into a single category, while more complex models \citep{NosofskyGluckPalmeriMckinleyGlauthier1994, NosofskyPalmeri1998} may use hierarchical categories to form a more comprehensive representation of examples. These models can vary wildly in both complexity and effectiveness, although in all cases they tend to act more as classifiers than models of human memory.

The models we have looked at so far do not offer a combination of the properties we want. Prototype-based models only store prototypes rather than individual states, have little in the way of meaningful update mechanics, and often lack biological plausibility. The Hopfield network is a much more interesting model, allowing for storage of individual states, interesting update processes, and biologically plausible learning rules. Furthermore, the Hopfield network already has some basis as a model of human memory, and hence prototype formation may have some psychological interpretation. Our work focuses on the emergence of prototypes in the Hopfield network, allowing for a natural progression from recall of individual states to recall of prototype states.

\subsection{Early Associative Memories}

More general models of associative memory have been studied extensively, much more than prototype models alone. Steinbuch introduces one of the earliest models of associative memory, the Learning Matrix \citep{Steinbuch1963, Steinbuch1965}. In this model, states are learned against binary key vectors with a single bit ``on''. The key vectors are also assumed to be orthogonal so that no two memories share the same row in the matrix, akin to a one-hot encoding. The interpretation set forth by Steinbuch is that the states represent the ``characteristics'' of a memory/item, while the key vector represents the memory/item ``meaning''. Steinbuch's work originally describes the implementation of the model in physical hardware, describing the learning process in terms of associating currents in a matrix of wires, but we will discuss the content in the modern context. 

Learning Matrices store states in an outer product process, like Hebbian learning, but the constraint of orthogonal key vectors assures no crosstalk between items. This allows for perfect recognition; presenting the network with an item's ``meaning'' vector will result in recognition of exactly the ``characteristics'' vector of that same item, as this is as simple as storing each item as a distinct row of a matrix. Although this model has perfect recognition it is not biologically plausible, requiring outside computation to generate key vectors, and without an iterative update process the dynamics of this model are non-existent. More importantly to our work, because the key vectors are kept intentionally orthogonal there is no chance for prototype formation. Two identical learned states would still occupy two separate rows of the weight matrix and do not influence one another during iteration. A collection of similar states can never create a single strong representative in Steinbuch's model by design.

Kohonen proposes another early associative memory, the Correlation Matrix Memory \citep{Kohonen1972}, which demonstrates many similarities to the Hopfield network. States are encoded in a matrix taken by the outer product of two vectors, although Kohonen allowed for heteroassociative learning, pairing a key state \(q\) with datum state \(x\):
\begin{align*}
    W &= \sum_\mu x^\mu \otimes q^\mu
\end{align*}
while the Hopfield network with Hebbian learning focuses on autoassociative learning, where \(q = x\). Kohonen also slightly alters the update method compared to the Hopfield network, in particular forgoing the use of a nonlinear function. The Correlation Matrix Memory update rule multiplies a probe (one of the stored keys, or a corrupted key) with the weight matrix, resulting in a final state related to the associated datum:
\begin{align*}
    x = Wq
\end{align*}
Due to the lack of a nonlinear function, Correlation Matrix Memories are linear in their storage and stability. If \((q^1, x^1), (r^2, y^2)\) are key data pairs stored in the model then any combination of the keys \((\alpha q^1 + \beta q^2)\) will recollect a combination of the data \((\alpha x^1 + \beta x^2)\) (at least, when the keys are orthogonal).

Kohonen discusses this model and other aspects of associative memory (with a focus on heteroassociative memories) in other work \citep{Kohonen1978} but the details are not overly relevant to our work due to the absence of any discussion of prototypes. However, it is clear from the above that Correlation Matrix Memories do not conform to our expected prototype behavior as although a probe may result in a combination of many related learned states, no particular combination is preferred. A linear model of associative memory cannot support prototype formation in a useful way, as there will never be a single strong representative when all combinations of learned states are stable.

\subsection{Statistical Physics and the Long Range Spin-Glass Model}
The Hopfield network has been studied extensively from the perspective of artificial intelligence, but the model has also received substantial attention from the statistical physics community. It has been shown to be entirely equivalent to the long-range, low-temperature spin-glass model with Ising spins \citep{Amit1985A, Mattis1985}, specifically the Sherrington-Kirkpatrick model \citep{KirkpatrickSherrington1978}. There is a great deal of research into bounds on network capacity and the exact form of stable states in the spin-glass model. Often the long-range spin-glass is investigated with a random Hamiltonian, equivalent to a random weight matrix in the Hopfield network, however (particularly after Hopfield's influential paper) the Hamiltonian matching that derived from Hebbian learning or other learning rules has also been studied. 

McEliece et al. find bounds on the capacity of the Hopfield network in the case of both random and orthogonal states \citep{McEliece1987}. For a network with \(N\) neurons, the bound on the absolute capacity (recognition with zero errors) of random states was found to be \(N / \left(4 \ln N\right)\) as \(N \rightarrow \infty\). Our work is less concerned about absolute capacity, or even relative capacity (recognition with some errors), as we are likely to push well above the network capacity to test strong prototype formation. We show that prototype formation is still likely even after the capacities given here are breached. Referring to Figure \ref{Fig:TaxonomyOfStates}, the proofs set forth by McEliece et al. are concerned with learned stable states while we are allowing many learned states to become unstable (be forgotten), allowing an unlearned state to form as a strong stable attractor representing those forgotten states. We place no restrictions on the number of states in the learned set, as long as the categories in the learned set are represented in the network's stable states. McEliece et al. also find an exponential increase in spurious attractors as the number of learned states exceeds the network capacity. This should be of concern to our work, as many spurious states may interrupt prototype formation, but empirically we find that these spurious attractors must form far away from our prototype formations, or at least any that form nearby are dominated by the prototype instead, as asymptotically all probes find the prototype attractors (see Section \ref{Section:ProportionOfProbesData}).

Amit, Gutfreund, and Sompolinsky investigate the spin-glass model and find the Hopfield network and another, more complicated model of memory proposed by Little \citep{Little1974}, are equivalent in the low-temperature limit \citep{Amit1985A}. Amit et al. use the Hebbian equivalent Hamiltonian in the limit of an infinite number of neurons, finding the \textit{only} stable states (up to degeneracy) are learned states. Amit et al. also investigate a non-zero temperature, finding spurious mixture states appear and become metastable. These mixture states are shown to have equal overlap with an arbitrary subset of learned states, and their number is shown to increase exponentially with the number of learned states. This is consistent with previous findings \citep{McEliece1987}. In contrast to spurious states arising from breaching the network capacity, the mixture states found here are due to an increased temperature, a hyperparameter that is rarely nonzero in the Hopfield network. Notably, as is a theme throughout the statistical physics research, the exact form of these mixture states is left very general. Without imposing additional structure on the learned states it is difficult to narrow down the form of the mixture states; the only claim is that mixture states are a combination of a subset of learned states. This result is the most similar to our work on prototype formation, as prototype states are almost certainly a combination of some learned states, since we are specifically interested in learned states that are highly correlated. However, our work builds on the forms presented in the series of papers from Amit et al. as we make predictions about states with \textit{specific} structure, rather than the more general case of \textit{any} mixture state. We also note that prototype states have distinct storage characteristics compared to mixture states \citep{Gorman2017}. Thankfully, this focus on the specific makes our analysis much less complex than Amit et al. as well.

Amit et al. continue their investigation into the spin-glass model, allowing the number of learned states to increase to infinity alongside the number of neurons, with the limiting ratio of \(\alpha = \frac{p}{N}\) for \(N\) neurons and \(p\) states \citep{Amit1985B}. In the low-temperature limit, and hence of particular interest to our work, they find a critical value of \(\alpha \approx 0.138\), above which no learned states are stable. Amit et al. also find an interesting phase of the spin-glass, the ferromagnetic phase, for roughly \(\alpha \lesssim 0.05\) in the low-temperature limit. Near the upper limit of this phase the learned states are the only stable states, up to degeneracy. This matches the experiments by Hopfield \citep{Hopfield1982}, and later derivations by Hertz \citep{Hertz1991}. Amit et al. also find the average ratio of errors to neurons in a learned state goes to zero only when \(\alpha\) goes to zero, but the \textit{total} number of errors in a state goes to zero with \(\alpha < \frac{1}{2 \ln N}\), as predicted by McEliece \citep{McEliece1987}. Still, no further constraints are placed on the form of the mixture states that appear, either from non-zero temperature or from exceeded capacity. 

Sompolinsky further extends this work by investigating the spin-glass model with an updated Hamiltonian \citep{Sompolinsky1986}, moving away from the Hebbian equivalent to one that saturates weights during learning. Weight saturation allows novel connections to create stronger weights compared to the Hebbian learning rule, i.e. states with many shared features will not dominate the weight matrix. The weights representing co-occurring features quickly saturate the corresponding weights, while other weights may only receive a handful of increments but will still meaningfully impact the network dynamics. Sompolinsky goes on to analyze the dynamics of this learning rule in great detail, finding for a small number of learned states the saturating learning rule can improve the recall. While our work will focus mostly on Hebbian learning, Sompolinsky's work here is of interest as a saturating learning rule would drastically change the storage of prototypes. It would be interesting to revisit saturating learning rules to see if they support prototype learning.

Amit, Gutfreund, and Sompolinsky return to study the dynamics of the Hopfield network under Hebbian learning as the network reaches the capacity of learned states \citep{Amit1987}. The phases of the spin-glass model are investigated, building on previous work. Of particular interest are the phase transitions, where network behavior changes dramatically, and mapping a comprehensive phase diagram. Amit et al. investigate the model across both \(\alpha = p/N\) and the temperature \(T\). Two phases are found that are useful as an associative memory with low temperature; \(\alpha \lesssim 0.05\) and \(0.05 \lesssim \alpha \lesssim 0.138\). For \(\alpha \lesssim 0.05\) the model exists in a ferromagnetic phase such that global energy minima are learned states. This means update methods employing limit process (such as simulated annealing) should always recall a learned state. For \(0.05 \lesssim \alpha \lesssim 0.138\) the model exists in a spin-glass phase, where learned states are still local minima, but not necessarily global minima, allowing for significantly reduced recall or perhaps only recognition. Spurious states may now make up global minima, specifically spin-glass states that are random and uncorrelated with the learned states. The model still operates as an associative memory as long as the probe states are sufficiently near the learned states. These phases extend to temperatures above \(T=0\), but we are particularly interested in the low-temperature limit. Interestingly, it is found the maximal value of \(\alpha \approx 0.138\) is not found at \(T=0\) but slightly higher, indicating a small non-zero temperature can extend the spin-glass region of the model allowing for larger capacities. The discussion on the form of mixture states in this work does not advance significantly from previous works.

\subsection{Alternative Learning Rules}

Diederich and Opper investigate the capacity of the spin-glass model by changing the learning rule away from Hebbian while keeping the biological plausibility of local learning \citep{Diederich1987}. The proposed learning rule is incremental, iterating over the learned set and checking if each state is stable in the network. Unstable states contribute their Hebbian matrix to the weight matrix. This rule shows some similarity to the error-correcting Hebbian, also known as the Delta rule or Widrow-Hoff rule \citep{WidrowHoff1960}, in that it applies updates only from unstable states. Notably, the delta rule only updates weights to the unstable \textit{neurons}, while Diederich and Opper's rule has unstable states contribute to the entire weight matrix no matter which neurons are stable. Diederich and Opper show their proposed learning rule will terminate if the states can be learned. For example, it has been shown that for a \(N \times N\) matrix, an upper limit of \(2N\) states can be stored. If we present Diederich and Opper's learning rule with a set of more than \(2N\) states, the learning rule will never converge. For this reason, we forgo analyzing Diederich and Opper's learning rule with respect to our work on prototypes, as we will often present far more than the upper limit on storage, and hence the rule will often not converge. 

The Storkey learning rule \citep{Storkey1997} similarly attempts to increase the capacity of the Hopfield network. Storkey introduces a local, incremental rule that does not require a potentially long limit process to achieve the final weight matrix. Instead, for each state the Hebbian contribution is added and a form of the pre-synaptic and post-synaptic fields are subtracted. Storkey shows this rule has a higher absolute capacity in the Hopfield network than the Hebbian, at least for random states and in the limit of a network with infinite neurons. Experimentally, the relative capacity is also higher than the Hopfield network with Hebbian learning. Like Sompolinsky's proposed learning rule above, it would be interesting to apply our work on prototypes to Storkey's rule, but we leave that for future work.

\subsection{Dense Associative Memories}

Recently, associative memories have had a major revival. Krotov and Hopfield's paper on a generalization of the Hopfield network \citep{KrotovHopfield2016} catalyzed renewed interest in the field, along with work relating the new models to attention mechanism \citep{HopfieldIsAllYouNeed2021}. The main generalization Krotov and Hopfield introduce is a non-linear activation function in the calculation of the energy. The traditional Hopfield network, with standard Hebbian learning, has energy function:
\begin{align*}
    E\left(\xi\right) &= - \frac{1}{2} \sum_{i,j} W_{ij}\xi_{i}\xi_{j} \\
    W_{ij} &= \sum_\mu \xi^\mu_i \xi^\mu_j
\end{align*}
and hence is quadratic in the states. The introduction of a non-linear function, the interaction function \(F_n\) with interaction vertex hyperparameter \(n\), allows Krotov and Hopfield to rewrite the energy with rectified polynomials:
\begin{align*}
    E\left(\xi\right) &= -\sum_\mu \sum_i F_n(\xi^\mu_i \xi_i) \\
    F_n(x) &= \begin{cases}
        x^n & x \geq 0 \\
        0 & x < 0
    \end{cases}
\end{align*}
which explicitly allows a non-quadratic energy. Krotov and Hopfield also introduce a slightly different update rule: instead of simple multiplication of a state by the weight matrix, the difference in energies between a neuron being ``on'' and ``off'' is calculated and is fed into an activation function. This update rule employs the same argument to demonstrate convergence as the traditional Hopfield network; at each step, the energy of the entire network must either remain constant or decrease, assuming a monotonically increasing activation function. While it is not immediately obvious, these generalizations do indeed reduce to the traditional Hopfield network for specific values, namely with hard limiting activation functions and interaction vertex \(n=2\) \citep{KrotovHopfield2016, Demircigil2017}. These alterations define the family of models known as Dense Associative Memories, or Modern Hopfield Networks.

Krotov and Hopfield investigate the effect of varying the interaction vertex. Small values of \(n\) give many learned states the ability to impact a probe, as even a small overlap will contribute significantly to the energy. Larger \(n\) will quickly diminish the contribution from learned states with small overlap, at least in comparison to states with large overlap. Because the energy function is polynomial, more specifically non-linear for all \(n\) of interest, low overlap contributions are diminished faster than high overlap ones, dramatically altering the behavior of the network. Krotov and Hopfield give examples of this, including a complete analysis of the solution of the XOR problem (in which it is shown \(n\geq2\) is required when using rectified polynomials), and an experiment with a slightly modified DAM on the MNIST dataset \citep{MNIST}. The modified DAM had a set of neurons, corresponding to the data/pixels, clamped to never relax, while the remaining neurons, corresponding to one hot encoded class labels, were allowed to relax with a hyperbolic tangent activation. These modifications are a step away from a pure associative memory, although it is shown the modified network is equivalent to a three-layer feed-forward neural network (another major result of the paper). A notable decrease in training time was observed on the MNIST dataset between \(n=2\) and \(n=3\). A similar training speed-up is observed when deep feed-forward networks move from hyperbolic tangent and sigmoid activations to rectified linear units \citep{Krizhevsky2012}, the equivalent of moving from \(n=1\) to \(n=2\) in the DAM. 

Most interesting to our work are the observations Krotov and Hopfield made on the feature-to-prototype transition. For small \(n\), DAMs appear to operate as feature matchers; learned states are decomposed into their common features which are stored and later compared against new probes to guide recall. This regime dominates for small values of the interaction vertex, as many learned states can make significant contributions to the network dynamics across all of state space. As \(n\) increases, only nearby learned states contribute to probe iteration. Krotov and Hopfield showed empirically that for large values of the interaction vertex prototypes were stored in the network, with new probes influenced by a single prototype rather than a large number of features. Interestingly, this did not increase the error on the dataset considerably (backed by later findings \citep{KrotovHopfield2018}), although MNIST is not a particularly challenging dataset. 

In our work, we also find prototype behavior in the Hopfield network, although we focus on the traditional Hopfield network. It is here we find a distinction between our work and Krotov and Hopfield's, namely that we show prototype behavior for low \(n\) in the DAM context (exactly \(n=2\)). Krotov and Hopfield find only feature-matching behavior at this value, and our work shows a novel and interesting contrast to this. Secondarily, Krotov and Hopfield use backpropagation to train the DAM in their experiments, starting with a set of random ``memories'' \(\xi^{\mu}\) that are themselves updated during training. We are interested in using examples from the dataset to recover prototype behavior using learning rules that may better represent learning in the brain. Storkey \citep{Storkey1997} notes three important properties of a learning rule; locality (weight updates only depend on information from the pre-synaptic and post-synaptic neurons), incrementality (weight updates can be applied to an existing network to learn new states, without referring to old states), and immediacy (the weight update process finishes in a fixed time, not as part of a limit process or potentially non-terminating algorithm). Our work focuses on Hebbian learning; more biologically plausible than backpropagation in all three areas set by Storkey. Finding prototype representation in a more biologically plausible context may make for more applicable work in the fields of psychology and neuroscience.

DAMs have not remained untouched since the initial paper. Krotov and Hopfield continue their work by looking at adversarial examples in Dense Associative Memories \citep{KrotovHopfield2018}. In one experiment, adversarial examples are generated by initializing a probe to random noise, then employing gradient descent to alter the probe to a minima in the objective function, i.e. a probe the network would classify with extreme confidence. Krotov and Hopfield performed this experiment again over the MNIST dataset and found for small values of the interaction vertex the generated examples match those found in deep neural networks; speckled, still very random images that are classified very strongly. As \(n\) grows, the images generated take on more structure, until eventually all examples have semantic meaning (\(n \approx 30\)), i.e. the generated images were identifiable by a human as an example of the class. In another experiment, adversarial examples were constructed by taking images from the dataset and using gradient descent to move the image to be placed exactly on a decision boundary where the model is least confident about a classification. This is the inverse of the above, in deep neural networks these examples have high semantic meaning but are classified with very low confidence (often incorrectly or randomly) by the network. Krotov and Hopfield find the typical examples for small values of the interaction vertex, but again as \(n\) increases these examples shift such that they correspond to mixtures of the prototypical images on either side of a decision boundary (e.g. an image with a 4 and 8 overlaid). Finally, experiments are performed in which examples are created in one network and transplanted to another. Examples generated in networks with small interaction vertices were found to sometimes fool other networks with small (but different) interaction vertices but did not fool networks with larger interaction vertices. This work employs some of the prototype behavior seen in the initial paper, with networks in the prototype regime (large \(n\)) appearing to be more resistant to adversarial examples than those in the feature-matching regime. It should be noted the high \(n\) networks were still able to be fooled by adversarial examples generated by that network, but are resistant against transferred examples from lower \(n\) networks. This work could well be expanded by investigating the link between adversarial example resistance and the increasing interaction vertex, keeping in mind Krotov and Hopfield's previous work relating the increasing interaction vertex to prototype formation in DAMs.

Others have also looked into the Dense Associative Memory. Demircigil et al. \citep{Demircigil2017} formalize many of the claims Krotov and Hopfield made around increased storage capacity \citep{KrotovHopfield2016}. The proofs employ a network with slightly altered dynamics; rather than taking the difference in energies as proposed by Krotov and Hopfield, neurons are updated based on a variation of the Sherrington-Kirkpatrick spin-glass with higher dimensional spins \citep{Bovier2001}. Demircigil proves Dense Associative Memories can store exponentially many (in the number of neurons) states in both absolute and relative capacity, although for rectified polynomial interaction functions with finite \(n\) the basins of attraction may shrink and vanish. Allowing the interaction vertex to grow to infinity, i.e. replacing the rectified polynomial with the exponential function \(F(x) = e^{x}\), Demircigil surprisingly finds the basins of attraction of these learned states remains almost as large as in the traditional Hopfield network. Demircigil shows this by proving a probe near a learned state will diverge with probability approaching zero, setting bounds on ``nearby'' based on the number of learned states.

Demircigil's work is extended by replacing the energy function with a complicated form based on the log-sum-exponential function, allowing for continuous states, and replacing the update rule with the softmax function \citep{HopfieldIsAllYouNeed2021}. Among a variety of results, Ramsauer et al. show their version of the Hopfield network retains exponential storage capacity in the number of neurons and will converge to a learned state in a single step. Ramsauer et al. also construct their Hopfield network to be differentiable and hence integrable into deep neural networks, and have shown their network has many similarities to the attention mechanism \citep{Bahdanau2014, Vaswani2017} used in transformer architectures. More relevant to our work, these exciting developments in the fields of associative memory have made far strides from the traditional Hopfield network, where we make claim to a novel discovery without altering the original architecture.

\subsection{Other Works on Prototypes}

Recent works attempt to distinguish between learned, spurious, and prototype states in the traditional Hopfield network \citep{Gorman2017}. While distinguishing between the former two classes has been of interest for many years \citep{Robins2004}, and has many applications \citep{Athithan1997, Robins1998}, the addition of the prototype class is novel in the study of distinguishing states. Gorman et al. share our definition of prototype: an unlearned stable state that represents some subset of learned states. Gorman et al. find there is a class of states significantly different from both learned states and other spurious attractors when analyzing the stability ratio (the ratio of the energies of the most and least stable neurons for a state). This class appears similar to prototype states, although the existence of prototypes is taken as an assumption. Our work proves the assumption, showing prototypes do indeed form for data of the form studied by Gorman et al.

There have also been several papers mentioning prototypes that either use different terminology (e.g. using ``prototype'' to mean learned state) or only mention prototypes in passing. Wu and Batalama implement an associative memory as a two-layer dense feed-forward neural network \citep{Wu2000}, with the hidden layer having one neuron for each ``prototype'' state stored. In this case, it appears that ``prototype'' is synonymous with learned state rather than psychological prototypes. Gascuel et al. also use ``prototype'' in this manner in a study on distinguishing prototype (learned states) and parasite (spurious states) attractors \citep{Gascuel1994}. Gascuel et al. appends a checksum code to the end of each state to be stored, which allows for each recalled state to be checked against its recalled checksum meaning spurious states have to also randomly stabilize with the correct checksum, which naively has a probability of \(2^{-d}\) for checksum of \(d\) bits. 

Berlemont and Nadal introduce a variation of Hebbian learning, one based on confidence, to aid in determining category membership \citep{Berlemont2022}. By only updating the weights on examples for which the network is not confident, performance on the task is improved compared to similar models trained using reward-based Hebbian learning. While category membership may appear to make reference to prototype-category theory and hence psychological prototypes, it instead refers to something closer to the notion of classification.

Bauckhage, Ramamurthy, and Sifa develop a method of determining prototypes of a dataset using Hopfield networks \citep{Bauckhage2020} using the power of optimization. By restating the problem as a quadratic unconstrained boundary optimization, it is possible to solve this problem by using this form as the energy function of a traditional Hopfield network as shown by Hopfield and Tank \citep{HopfieldTank1985}. Bauckhage et al. present an interesting way to extract prototypes as vector quantization in a dataset, but the methods are tremendously different from our work which looks at learning prototype representations directly from a dataset of binary states.
\section{Formalization of the Hopfield Network} \label{Section:Formalization}

The Hopfield network is a simple model, but still requires rigorous formalization for our theory. The network weights are a square matrix \(\boldsymbol{W} \in \mathbb{R}^{N \times N}\), for network dimension \(N\). The weight matrix is typically symmetrical, although some applications allow for an asymmetrical matrix; trading guaranteed iteration convergence for a larger capacity \citep{Ma1999}. The weight matrix also typically enforces diagonal entries of zero. For our theory we consider the symmetric matrix with zero diagonal. States are binary vectors from \(\{-1, 1\}^N\). For a state \(\xi\) at time \(t\), the \(i^{\text{th}}\) neuron is updated by:
\begin{align*} 
    \begin{split}
        \xi_i(t+1) &= \sigma\left(\left(\boldsymbol{W}\xi(t)\right)_i\right) \\
            &= \sigma\left(\sum_j W_{ji} \xi_j(t)\right)
    \end{split}
\end{align*}

Where \(\sigma\) is the hard limiting activation function keeping the updated state \(\xi(t)\) in the domain \(\{-1, 1\}^N\).

\begin{align*}
    \sigma(x) = 
    \begin{cases}
        1 &\text{if } x\geq0 \\
        -1 &\text{if } x<0 
    \end{cases}
\end{align*}

The state's time dependence, denoted by \(t\), is excluded in all further equations to avoid clutter.

Note that indices are updated asynchronously. Synchronous update methods exist but are uncommon. It has been shown that asynchronous updates with a symmetric weight matrix guarantee convergence in finite time \citep{Hopfield1982, Bruck1990}.

\subsection{The Hebbian Learning Rule} \label{Section:HebbianLearningRule}

The Hebbian learning rule is the simplest method to learn the weight matrix \(\boldsymbol{W}\). Given a set of learned states of size \(K\):
\begin{align*}
    \boldsymbol\xi = \{\xi^1, \xi^2, \dots, \xi^K\}
\end{align*}

The Hebbian learning rule is given by:
\begin{align} \label{Eqn:HebbianLearningRule}
    \begin{split}
        W_{ii} &= 0 \\
        W_{ji} &= \frac{1}{K} \sum_k \xi^k_j \xi^k_i
    \end{split}
\end{align}

The Hebbian learning rule in the Hopfield network has been analyzed in great detail \citep[summaries ][]{Hertz1991, Haykin2009}. Importantly the Hebbian learning rule will act to stabilize only the learned states from \(\boldsymbol\xi\). No other states are intentionally stabilized, but spurious states can and do form. We will show prototype formation in the Hopfield network under Hebbian learning, although prototypes can and do form under other learning rules. The Hebbian learning rule is particularly nice to analyze due to a lack of stochastic operations and limit processes.

\section{Analysis of Prototype Formation} \label{Section:PrototypeFormationAnalysis}

The learning process typically produces a few well studied states in the Hopfield network, shown in Figure \ref{Fig:TaxonomyOfStates}. We are familiar with learned states stabilizing as attractors, as well as spurious states in their multiple forms. Learning may also cause another class of states to emerge. Under specific conditions, prototype states may stabilize that have unique properties in the Hopfield network. We show the Hebbian learning process causes large sets of correlated states to make stabilizing contributions to a single representative state. We then argue this state is likely to have prototype properties.

In this section we start with a set of vectors, outside the Hopfield network. We discuss mixture vectors of a set and construct the maximally representative mixture vector, or simply the representative vector. We then use this set in the Hebbian learning rule from Equation \ref{Eqn:HebbianLearningRule} to obtain an expression for the weight matrix in terms of the representative vector. We make the set of vectors the learned states of the Hopfield network and show the representative vector has prototype properties depending on the original set. For a set of vectors that is large and highly correlated we find the Hebbian learning rule stabilizes the representative vector more than any individual vector, therefore dominating the attractor space in place of the learned states.

\subsection{Defining a Representative Vector} 
\label{Section:DefineRepresentativeVector}

For an arbitrary set of binary vectors over \(\{-1, 1\}^N\), a mixture vector of the set is another binary vector over \(\{-1, 1\}^N\). Mixture vectors are simply the result of choosing values from among the set for each index. For any index \(i\) a mixture vector may have any value an item takes at index \(i\). Therefore, all set items are trivially mixture vectors, and mixture vectors may cover the entire domain if the set allows. Mixture vectors are too broad a class to be useful, so we define a particular mixture vector -- the maximally representative mixture vector (or simply representative vector) -- by only taking the most common value at each index.


Given a set of vectors of size \(K\)
\begin{align*}
    \overline{\xi} = \{\xi^1, \xi^2, \dots, \xi^K\}
\end{align*}
We construct the representative vector:
\begin{align} \label{Eqn:RepresentativeState}
    \begin{split}
        \psi &\in \{-1, 1\}^N \\
        \psi_i &= \text{sign}\left(\sum_{k} \xi^k_i\right)
    \end{split}
\end{align}
As our vectors are binary, the sign of the sum will give us the most common element from among the set. If there are an equal count of values at an index \(i\) we arbitrarily define \(\psi_i\) as either \(-1\) or \(1\). We later we find the prototype behavior is equivalent given either choice.

We can now rewrite each vector \(\xi^k \in \xi\) as:
\begin{align} \label{Eqn:rewriteStateAsRepresentative}
    \begin{split}
        \xi^k &= \psi (1 - 2\delta^k) \\
        \delta^k &\in \{0, 1\}^N
    \end{split}
\end{align}
The new vector \(\delta^k\) has value \(0\) at the indices where the vector \(\xi^k\) agrees with the representative vectore and value \(1\) where the vector disagrees. This implies each \(\delta^k\) is a vector of Bernoulli distributed variables, with each index \(i\) being drawn from a separate distribution with Bernoulli parameter \(p_i\). The Bernoulli parameters measure the agreement with the representative vector. Small values indicate high agreement, meaning the set \(\overline{\xi}\) is highly correlated / low noise. Large values indicate low correlation / high noise. This becomes important in our prototype analysis, as highly correlated sets should produce strong prototypes, so we expect some reliance on the Bernoulli parameters.

 Note that all \(p_i\) must lie in the range \([0, 0.5]\) instead of the typical \([0, 1]\). The lower bound corresponds to the situation where all vectors agree with the representative at an index: \(\forall k: \delta^k_i = 0\) hence \(p_i=0\). The upper bound arises from our construction. If an index \(i\) has \(p_i > 0.5\) this implies more than half the vectors in \(\overline{\xi}\) disagree with the representative vector. This contradicts our construction of \(\psi\) which has the majority of vectors in \(\overline{\xi}\) agree by Equation \ref{Eqn:RepresentativeState}. The maximal value of \(p_i\) occurs where exactly half of the set \(\overline{\xi}\) has a value of \(-1\) at index \(i\) and half has value \(1\), exactly the situation where we have to arbitrarily choose a value, and either choice will result in \(p_i = 0.5\).

\subsection{Hebbian Learning and Representative Vectors} \label{Section:RepresentativeStateHebbianLearning}

The Hebbian learning rule attempts to stabilize a vector \(\xi\) by adding factors of \(\xi_j \xi_i\) to \(W_{ji}\). We will apply the Hebbian learning rule to a set of vectors \(\Xi\). In order to investigate the possibility of prototype formation for only a subset of vectors (e.g. some vectors do not belong to the class, or multiple prototypes are present) we select a subset \(\overline{\xi} \subseteq \Xi\) with representative vector \(\psi\). The remaining vectors will be separated into a term that we can analyze independent of the selected subset and its representative vector. Using Equations \ref{Eqn:RepresentativeState} and \ref{Eqn:rewriteStateAsRepresentative}, we will obtain an expression for the weight matrix in terms of the representative vector and the Bernoulli parameters. We will then interpret this result in terms of the stability of the representative vector.

Starting with the Hebbian learning rule, we separate the sum into factors based on membership of the chosen subset \(\overline{\xi}\). These vectors can then be rewritten using Equation \ref{Eqn:rewriteStateAsRepresentative}:
\begin{equation} \label{Eqn:HebbianWeightWorking}
    \begin{aligned}
        W_{ji} &= \frac{1}{K} \sum_k \xi^k_j \xi^k_i \\
        &= \frac{1}{K}\sum_{k \in \overline{\xi}} \xi^k_j \xi^k_i + \frac{1}{K} \sum_{k^\prime \notin \overline{\xi}} \xi^{k^\prime}_j \xi^{k^\prime}_i \\
        &= \frac{1}{K}\sum_{k \in \overline{\xi}} (\psi_j (1 - 2\delta^k_j)) (\psi_i (1 - 2\delta^k_i)) + \frac{1}{K} \sum_{k^\prime \notin \overline{\xi}} \xi^{k^\prime}_j \xi^{k^\prime}_i \\ 
        &= \frac{1}{K}\sum_{k \in \overline{\xi}} (\psi_j \psi_i) (1 - 2\delta^k_j) (1 - 2\delta^k_i) + \frac{1}{K} \sum_{k^\prime \notin \overline{\xi}} \xi^{k^\prime}_j \xi^{k^\prime}_i \\ 
        &= \frac{1}{K}\sum_{k \in \overline{\xi}} (\psi_j \psi_i) (1 - 2\delta^k_j - 2\delta^k_i + 4\delta^k_j\delta^k_i) + \frac{1}{K} \sum_{k^\prime \notin \overline{\xi}} \xi^{k^\prime}_j \xi^{k^\prime}_i \\ 
    \end{aligned}
\end{equation}

We have found a contribution of \(\psi_j \psi_i\) in the first sum. This looks like the stabilization term for the representative vector, i.e. we have shown the representative vector receives some stabilization from the selected subset. The additional factor \((1 - 2\delta^k_j - 2\delta^k_i + 4\delta^k_j\delta^k_i)\) can be simplified if we recall that each \(\delta\) is Bernoulli distributed: if the size of the subset is large enough we can approximate the sum with the expected value of the Bernoulli distribution multiplied by the number of random variables. Thankfully, the expected value of the Bernoulli distribution is a simple expression:
\begin{align} \label{Eqn:BernoulliExpectation}
    \mathbb{E}\left(X_{\text{Bernoulli}}\right) = p
\end{align}

We replace each \(\delta\) from the above with the Bernoulli parameter for that index:
\begin{align} \label{Eqn:GeneralHebbianRepresentativeStateWeight}
    W_{ji} &= \frac{|\overline{\xi}|}{K} (\psi_j \psi_i) (1 - 2p_j - 2p_i + 4p_jp_i) + \frac{1}{K} \sum_{k^\prime \notin \overline{\xi}} \xi^{k^\prime}_j \xi^{k^\prime}_i
\end{align}

Equation \ref{Eqn:GeneralHebbianRepresentativeStateWeight} is as far as we can take the general weight expression. If we also assume that \(\forall i,j: p_i = p_j = p\), meaning disagreements with the representative vector are equally likely across all indices, we can further reduce to:
\begin{align} \label{Eqn:SimpleHebbianRepresentativeStateWeight}
    W_{ji} &= \frac{|\overline{\xi}|}{K} (\psi_j \psi_i) (1 - 4p + 4p^2) + \frac{1}{K} \sum_{k^\prime \notin \overline{\xi}} \xi^{k^\prime}_j \xi^{k^\prime}_i
\end{align}

In the first term we find the expression for presenting the representative vector itself \(|\overline{\xi}|\) times multiplied by a factor measuring the agreement of the set with the representative, \((1 - 4p + 4p^2)\). If the examples agree very strongly with the representative vector, meaning \(p\approx0\), this additional factor will be large, close to 1 and the representative will be strongly stabilized. If the agreement is very weak, \(p\approx0.5\), this factor will be close to 0 and the representative is hardly stabilized at all. Therefore, subsets with low correlation (which result in large Bernoulli parameters) will likely not stabilize their representatives and are unlikely to form prototypes. This is the key that allows us to consider arbitrary subsets, despite the fact there are an exponential number of possible subsets. Although every subset has a representative vector, only the most correlated subsets will stabilize their representative vector with high probability.

\subsection{Stability of Representative States}

Applying our results so far to the Hopfield network, we move from abstract vectors to states -- actual patterns of activation within the network. Our set of vectors \(\Xi\) above become our learned states, and we continue to focus on the subset \(\overline{\xi}\) to investigate the representative vector, now representative state. We now have an expression for the weight matrix rewritten in terms of the representative state as well as the remaining unselected learned states. Next, we assess the stability of the representative state based on the weight matrix using the energy of the Hopfield network. The energy of a single index \(i\) of state \(\xi\) is defined as:

\begin{align}
    E(\xi)_i = \xi_i \sum_j W_{ji} \xi_j
\end{align}
The energy of a unit has a nice interpretation for stability: when the energy of index \(i\) is negative that unit is stable, and applying the update rule to it will not change the value of that unit. Conversely, an index with positive energy is unstable and \textit{will} change its value if the update rule is applied to it. 

A state \(\xi\) is stable if \(E(\xi_i) < 0\) for every index:
\begin{align} \label{Eqn:StabilityCondition}
    0 < \xi_i \sum_j W_{ji} \xi_j \qquad \forall i\in\{1, \dots, N\}
\end{align}

Using the stability condition in Equation \ref{Eqn:StabilityCondition} we link stability to the weight matrix and can assess representative state stability using the weight expression derived above. We replace \(W_{ji}\) in Equation \ref{Eqn:StabilityCondition} with the expression derived in Equation \ref{Eqn:SimpleHebbianRepresentativeStateWeight}, and replace the learned state \(\xi\) with \(\psi\) to find the stability condition of the representative state under Hebbian learning:

\begin{align*}
    0 &< \psi_i \sum_j W_{ji} \psi_j \\
        &= \psi_i \sum_j \left(\frac{|\overline{\xi}|}{K} (\psi_j \psi_i) (1 - 4p + 4p^2) + \frac{1}{K} \sum_{k^\prime \notin \overline{\xi}} \xi^{k^\prime}_j \xi^{k^\prime}_i \right) \psi_j \\
        &= \psi_i \frac{|\overline{\xi}|}{K} \sum_j (\psi_j \psi_i) (1 - 4p + 4p^2) \psi_j  \\
        &\qquad+ \psi_i \frac{1}{K} \sum_j\sum_{k^\prime \notin \overline{\xi}} \xi^{k^\prime}_j \xi^{k^\prime}_i \psi_j  \\
        &= \frac{|\overline{\xi}|}{K} \sum_j (\psi_i)^2 (\psi_j)^2(1 - 4p + 4p^2) \\
        &\qquad+ \psi_i \frac{1}{K} \sum_j \sum_{k^\prime \notin \overline{\xi}} \xi^{k^\prime}_j \xi^{k^\prime}_i \psi_j  \\
        &= \frac{N|\overline{\xi}|}{K} (1 - 4p + 4p^2) + \psi_i \frac{1}{K} \sum_j \sum_{k^\prime \notin \overline{\xi}} \xi^{k^\prime}_j \xi^{k^\prime}_i \psi_j  \\
\end{align*}
In the final step we have removed the factors of \(\psi_i^2 = 1\), as states are drawn from \(\{-1,1\}^N\). We have also replaced the sum over indices \(\sum_j\) with the constant factor \(N\), as the summand has no dependence on the sum.

\begin{align} \label{Eqn:StabilityConditionOfRepresentativeState}
    \forall i\in\{1, \dots, N\},\quad -|\overline{\xi}| (1 - 4p + 4p^2) < \psi_i \frac{1}{N} \sum_j \sum_{k^\prime \notin \overline{\xi}} \xi^{k^\prime}_j \xi^{k^\prime}_i \psi_j
\end{align}

Equation \ref{Eqn:StabilityConditionOfRepresentativeState} is a generalized form of the analysis of stability from Hertz \citep{Hertz1991}. Following Hertz's work we can derive a probability that the representative state is stable. We first assume that both \(K\) and \(N\) are large and that the remaining states not selected in \(\overline{\xi}\) are randomly distributed. This assumption will not always be valid, but it will suffice for an approximation. Hertz uses these assumptions to approximate the right-hand side of Equation \ref{Eqn:StabilityConditionOfRepresentativeState} as \(\frac{1}{N}\) times the sum of \(NK\) random Bernoulli trials. In our case, the right-hand side is the sum of \((K-|\overline{\xi}|)\) random trials as some states are in our selected subset, but we will take \((K-|\overline{\xi}|) \approx K\) to simplify.

The sum of these random variables forms a binomial distribution with \(\mu=0\) and \(\sigma^2 = \frac{K}{N}\). If \(N, K\) are both large, we approximate the binomial distribution with a normal distribution with the same \(\mu, \sigma^2\) by the central limit theorem. We can now formulate the probability of error in any index of the representative state:

\begin{align*}
    P_{\text{error}}\left(\psi\right) &= P\left(-|\overline{\xi}| (1 - 4p + 4p^2) < \psi_i \frac{1}{N} \sum_j \sum_{k^\prime \notin \overline{\xi}} \xi^{k^\prime}_j \xi^{k^\prime}_i \psi_j\right) \\
    &= \frac{1}{\sqrt{2\pi\sigma^2}} \int_{-\infty}^{-|\overline{\xi}| (1 - 4p + 4p^2)} \exp\left(\frac{-x^2}{2\sigma^2}\right) dx \\
    &= \frac{1}{2}\left(1+\erf\left(\frac{-|\overline{\xi}| (1 - 4p + 4p^2)}{\sqrt{2\sigma^2}}\right)\right) \\
    &= \frac{1}{2}\left(1-\erf\left(|\overline{\xi}| (1 - 4p + 4p^2)\sqrt{\frac{N}{2K}}\right)\right) \numberequation \label{Eqn:GeneralProbabilityOfRepresentativeStateStability}
\end{align*}

In the degenerate case where we choose exactly one state for our subset, we effectively set \(|\overline{\xi}|=1, p=0\), from which Equation \ref{Eqn:GeneralProbabilityOfRepresentativeStateStability} reduces to the more familiar:

\begin{align} \label{Eqn:ProbabilityOfStateStability}
    P_{\text{error}} = \frac{1}{2}\left(1-\erf\left(\sqrt{\frac{N}{2K}}\right)\right)
\end{align}
i.e. the exact expression Hertz derives.

Hertz calculates the maximum ratio of \(\frac{K}{N}\) that gives an acceptable probability of error. For example, in the case that the acceptable probability is \(P_{\text{error}}=0.0036\) Hertz calculates a maximum ratio of \(\frac{K_{max}}{N} = 0.138\), matching other theory and experiments.

\begin{figure}[H]
    \begin{center}
        \includegraphics[width=0.9\textwidth]{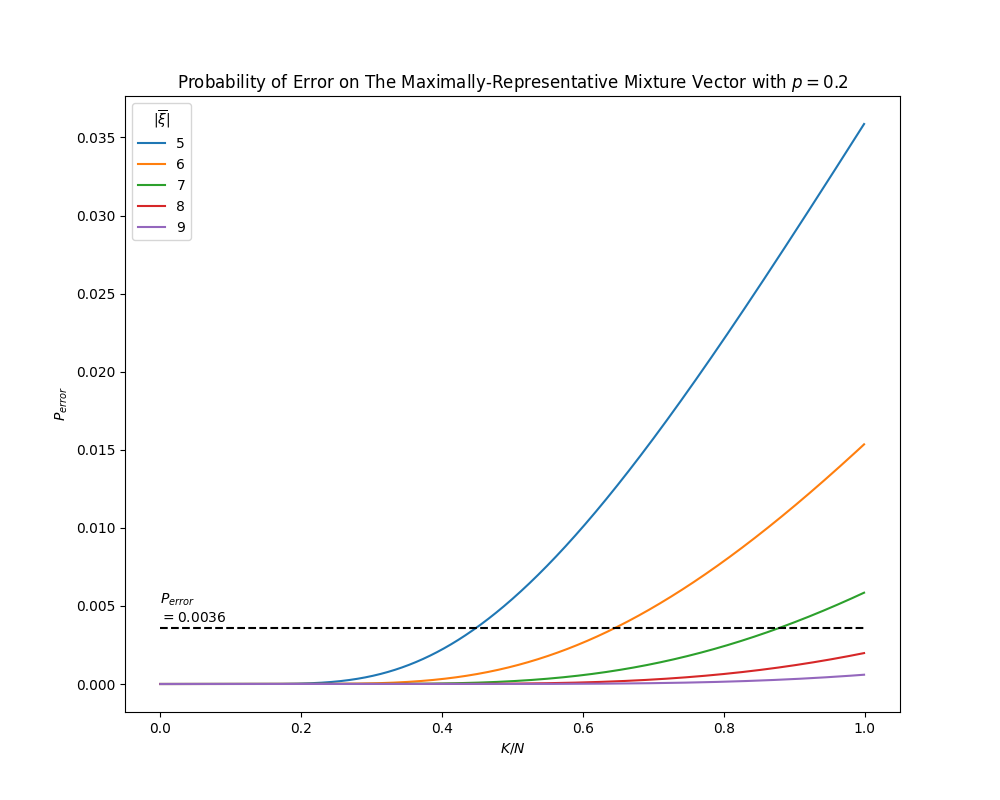}
    \end{center}
    \caption{Probability of Error of representative mixture states over ratios of K to N, holding the Bernoulli parameter. The horizontal line at \(P_{\text{error}}=0.0036\) is added to indicate the allowable \(P_{\text{error}}\) according to Hertz.}
    \label{Fig:pErrorHueEta}
\end{figure}

\begin{figure}[H] 
    \begin{center}
        \includegraphics[width=0.9\textwidth]{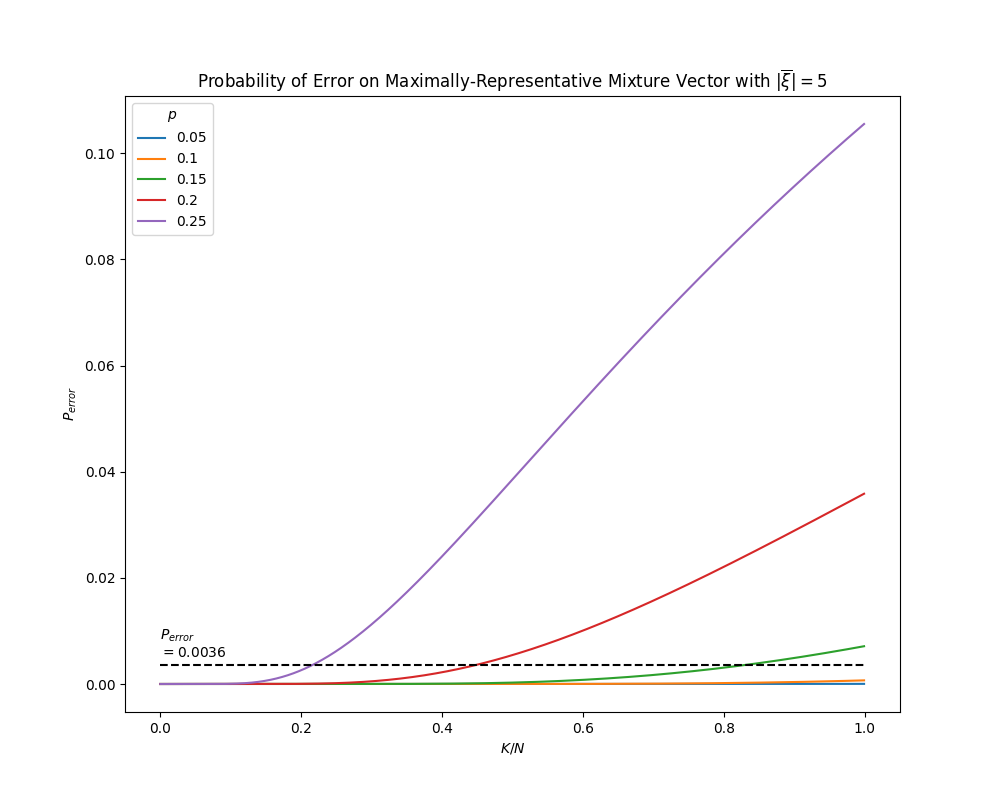}
    \end{center}
    \caption{Probability of Error of representative mixture states over ratios of K to N, holding the number of presented examples constant. The horizontal line at \(P_{\text{error}}=0.0036\) is added to indicate the allowable \(P_{\text{error}}\) according to Hertz.}
    \label{Fig:pErrorHueP}
\end{figure}

In Figure \ref{Fig:pErrorHueEta} and Figure \ref{Fig:pErrorHueP} we plot the value of \(P_{\text{error}}\) from Equation \ref{Eqn:GeneralProbabilityOfRepresentativeStateStability} with some select values of \(|\overline{\xi}|\) and the Bernoulli parameter. Each line shown is a slice through the higher dimensional space of all combinations of \(\frac{K}{N}\), \(|\overline{\xi}|\), and \(p\). We add to these plots the horizontal line at \(P_{\text{error}} = 0.0036\), the value of \(P_{\text{error}}\) Hertz used as a baseline to derive the capacity of the Hopfield network. Above this value of \(P_{\text{error}}\) the errors in a state tend to cascade and result in network degradation or total collapse of state stability. We see in Figure \ref{Fig:pErrorHueEta} that the probability of error on a representative state exceeds the critical value first for \(|\overline{\xi}|=5\), then \(|\overline{\xi}|=6\), and so on. This shows adding more example states makes the representative state more robust in the network, as expected. A similar result is seen in Figure \ref{Fig:pErrorHueP}, as a lower value of the Bernoulli parameter results in more robust representative state in the network. This analysis also forms the basis of determining the Hopfield network's capacity under Hebbian learning, and since we have derived identical results we can claim the network should have an identical capacity of \(\alpha \approx 0.138\) prototype states.

\subsection{On Representative States and Prototypes}

Throughout this section we have been analyzing the representative state of a set, which we took to be a subset of learned states in the Hopfield network. When our selected set consists of many highly correlated states they will almost certainly be forgotten as highly correlated states are difficult for the Hebbian process. By Equation \ref{Eqn:GeneralHebbianRepresentativeStateWeight} we expect the highly correlated set to make many strong stabilizing contributions to the representative state. While the learned states may be forgotten, if the representative state is stabilized enough (by number of examples and high agreement) it may still dominate the attractor space. This is shown empirically in Section \ref{Section:ProportionOfProbesData}. As the name suggests, we say the representative state represents the forgotten states in the attractor space, and has formed as a prototype. We also show experimentally that the representative state has distinct storage properties in the network. These properties match previous work on prototype formation in the Hopfield network \citep[see][]{Gorman2017} lending further credibility to representative states forming as prototypes. More traditional methods of identifying prototype states such as clustering of recalled states and identification of stability profiles are also explored in the following Section. In summary, we have shown the representative state of a large, highly correlated set is likely to form as a prototype under Hebbian learning.

\section{Empirical Analysis of Prototype Formation}

\begin{figure}[H] 
    \begin{center}
        \includegraphics[width=0.8\textwidth]{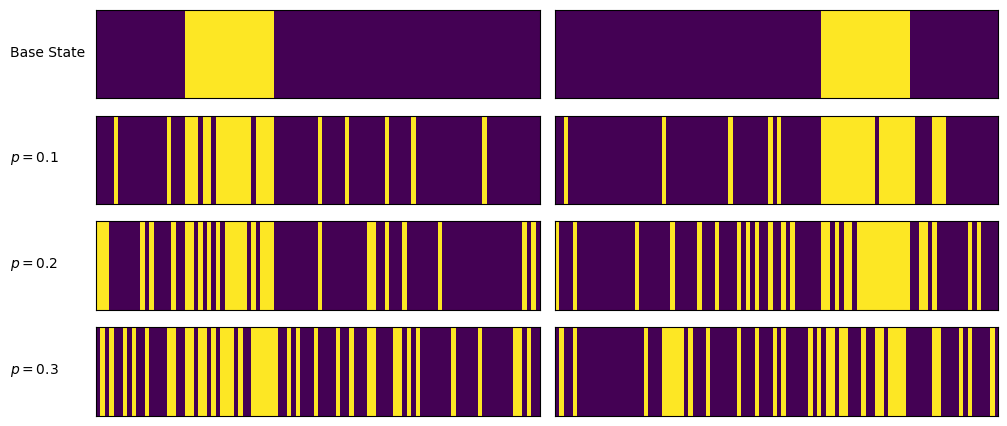}
    \end{center}
    \caption{Two representative states (top row) with examples constructed by applying increasingly more noise, corresponding to a larger Bernoulli parameter. A larger Bernoulli parameter results in states further from the representative. Yellow corresponds to \(1\), purple to \(-1\). Note the states are vectors, and the height shown here is for ease of visualization only.}
    \label{Fig:PrototypeStateExamples}
\end{figure}

We conduct experiments to corroborate the predictions made by our above theory. In our experiments we create and train on a dataset that should produce prototypes in the Hopfield network. We create such a dataset in the reverse of Section \ref{Section:DefineRepresentativeVector}. Starting with a random vector, we generate examples by inverting bits according to the Bernoulli distribution. The initial vector will be the representative vector of this set. Figure \ref{Fig:PrototypeStateExamples} shows the impact of the Bernoulli parameter on example generation. We then use the example vectors as a dataset with well known representative vector and Bernoulli parameter. Aggregating multiple such sets may allow for multiple prototype formation on the representative vectors as long as the crosstalk is minimal. This can be achieved by increasing the number of examples, decreasing the Bernoulli parameters / noise, as well as ensuring the representative vectors are far apart. Once we have a dataset, we can apply Hebbian learning and test for prototype formation. To test this, we must show a state represents a large set of learned states and dominates the nearby attractor space. We know our representative states already meet the former requirement by how we constructed the data, so as long as the representative states also dominate the attractor space we can claim they have formed as prototypes. We can show domination by probing the network many times and recording how many probes recall the representative states -- if many probes nearby those states recall them, they must not be inhibited by other attractors. Each experiment uses 100,000 probes total, generated again by applying Bernoulli noise to the original representative states.

Our experiments have the following structure. First, probe the network with states generated from our representative states. Find the most common final states from the probes. To see if our representative state has formed as the prototype over some other state, we measure the distance from the most recalled state to the nearest representative state. We expect this distance to be zero, but allow small deviations due to small example counts and crosstalk from other subsets. We also measure the proportion of probes that recall the most recalled state. If the distance is small and the proportion is large, we have found our representative state has formed as a prototype as predicted by our theory.

We conducted a grid search over many values of the network dimension, the number of prototypes, the number of examples for each prototype, and the Bernoulli parameter. Results of all of our experiments can be found in Appendix \ref{Apnx:AllData}.

\subsection{Prototype Formation with Additional Confounding States}

In our first experiment we measure prototype formation when training on a dataset containing a prototype subset as described above as well as additional uniformly random states. This is like having a traditional autoassociative memory task simultaneous with a prototype task, and allows us to see how these confounding states impact prototype formation. In this experiment, we learned only a single prototype in a network of 250 neurons.

\begin{figure}[H] 
    \begin{center}
        \includegraphics[width=0.9\textwidth]{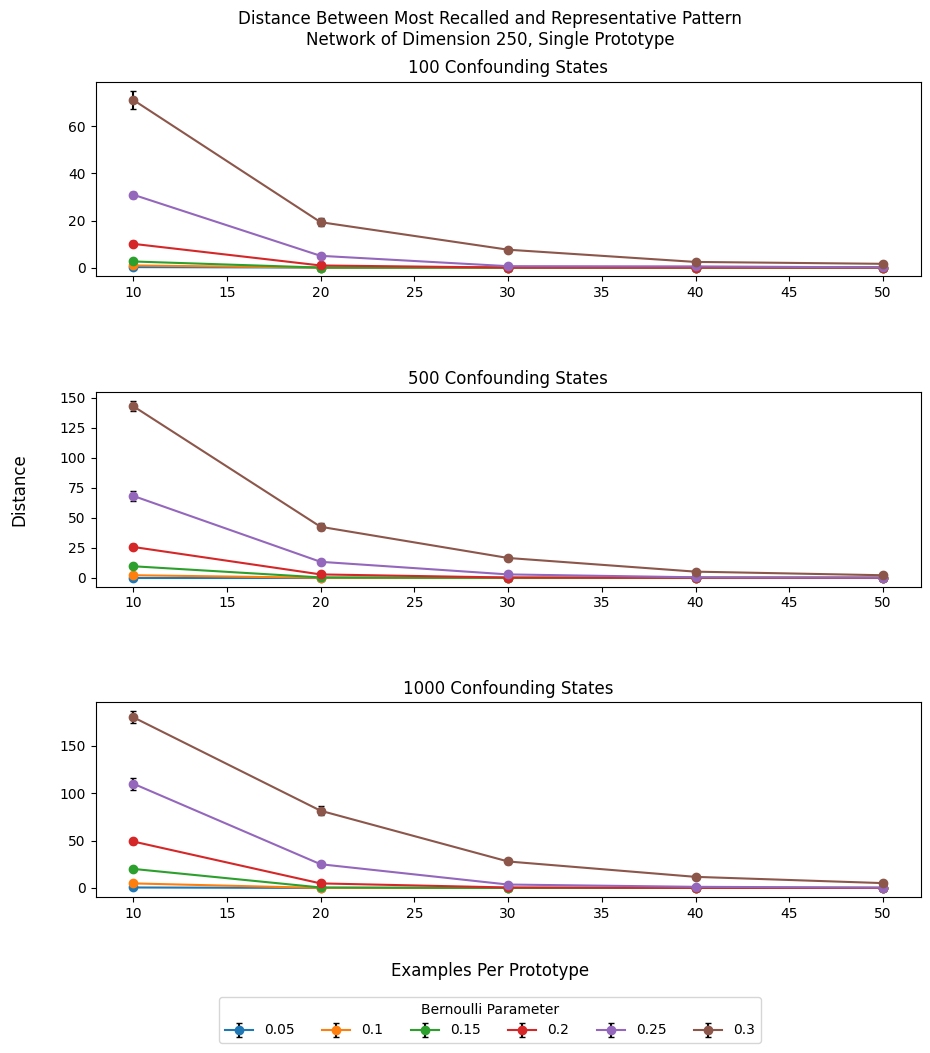}
    \end{center}
    \caption{Plot of the average Manhattan distance between the most recalled state and the representative state. Different plots correspond to different numbers of confounding states, while colors represent different Bernoulli parameters.}
    \label{Fig:ConfoundingMostRecalledDistance}
\end{figure}

\begin{figure}[H] 
    \begin{center}
        \includegraphics[width=0.9\textwidth]{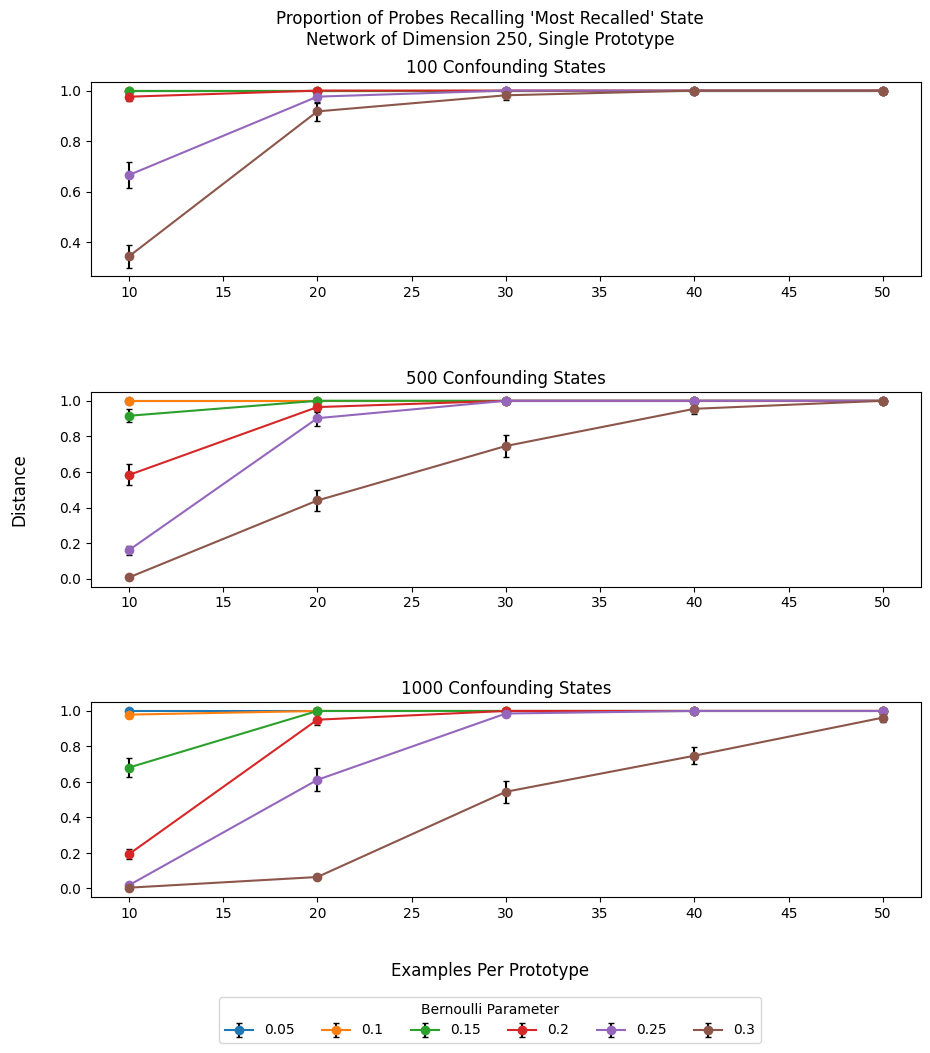}
    \end{center}
    \caption{Plot of the proportion of probes recalling the most recalled state. Different plots correspond to different numbers of confounding states, while colors represent different Bernoulli parameters.}
    \label{Fig:ConfoundingMostRecalledProportion}
\end{figure}

We see the distance increases as the Bernoulli parameter increases across all combinations of examples presented and confounding states. We also see the distance increases dramatically as the number of confounding states increases. At 1000 confounding states the network stabilizes an attractor far from the representative state for even a small amount of noise (e.g. Bernoulli parameter of \(0.15\)). However, in all situations, the distance to the representative state drops to near zero once the number of examples increases. We also see that the proportion of probes recalling the most recalled state increases towards one in the limit of many examples. Therefore, when the network is presented with sufficiently many examples we find the representative state is formed as a prototype, with strength influenced by the noise of the examples and the number of confounding states. These are the relationships we expect for a network presented with random states, and we demonstrate the Hopfield network forms prototype attractors even when the number of confounding states is very large relative to the prototype examples.

\subsection{Distance Between Most Recalled and Representative State} \label{Section:DistanceData}

In our remaining experiments we forgo confounding states, so every state in our training data is generated from a base representative state. We train the network on data with many representative states to see if multiple prototypes can form concurrently. Because of this, we take the top \(n\) most recalled states (for \(n\) representative states) and distances are measured from the most recalled state to the closest representative state. While this may seem like an information leak, as we must know beforehand how many prototypes we expect, we find there is a large disparity in the proportion of recalls between the top \(n\) states and the remainder when prototype formation occurs (see Section \ref{Section:ProportionOfProbesData}). We also make reference to the prototype capacity, denoted \(\alpha\), which measures the number of prototypes we expect the network to be able to store by the above theory. The prototype capacity matches the traditional capacity of the Hopfield network of \(\alpha \approx 0.138\).

\begin{figure}[H]
    \centering
    \begin{subfigure}{\textwidth}
        \centering
        \includegraphics[width=\textwidth]{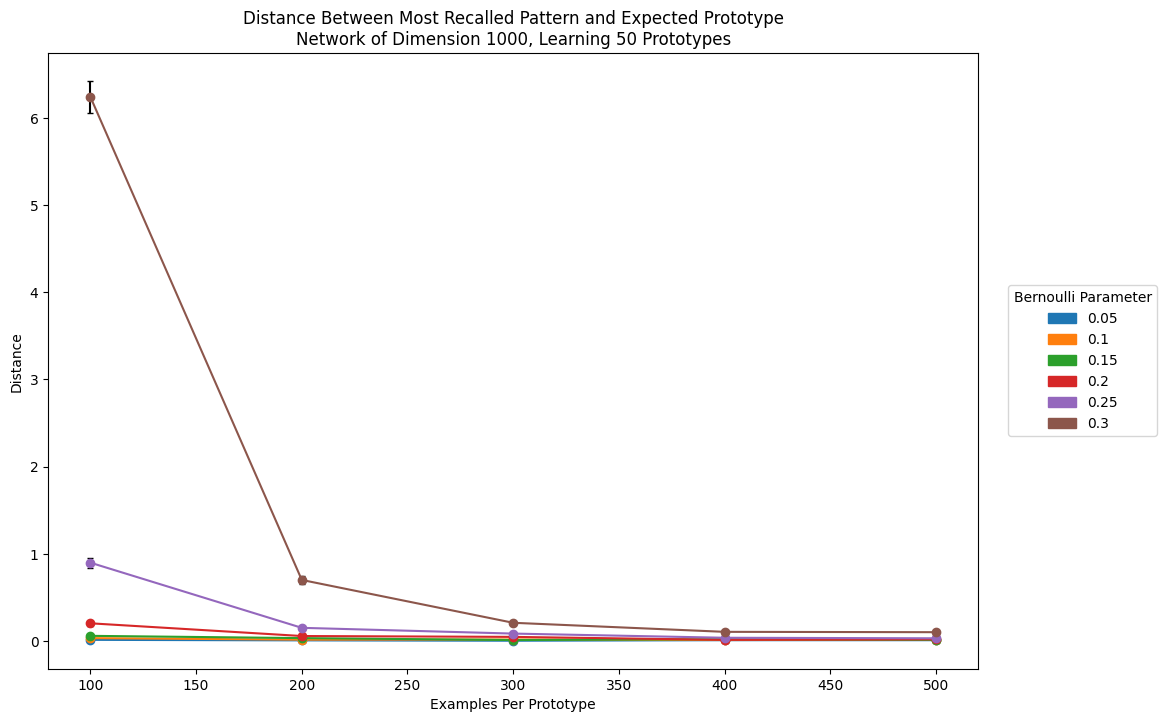}
        \caption{Below Prototype Capacity, \(\alpha = 0.05\)}
        \label{Fig:DistanceBelowCapacity}
    \end{subfigure}
    \hfill
    \begin{subfigure}{\textwidth}
        \centering
        \includegraphics[width=\textwidth]{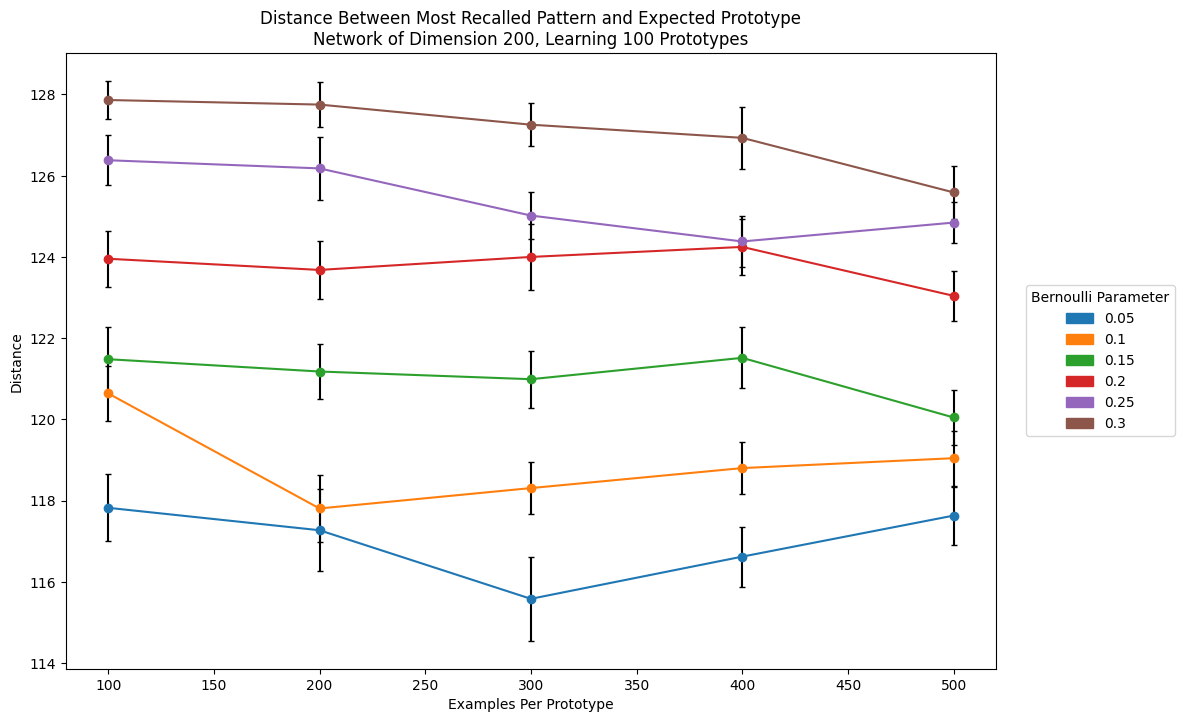}
        \caption{Above Prototype Capacity, \(\alpha = 0.5\)}
        \label{Fig:DistanceAboveCapacity}
    \end{subfigure}
\end{figure}
\begin{figure}[H]\ContinuedFloat
    \centering
    \begin{subfigure}{\textwidth}
        \centering
        \includegraphics[width=\textwidth]{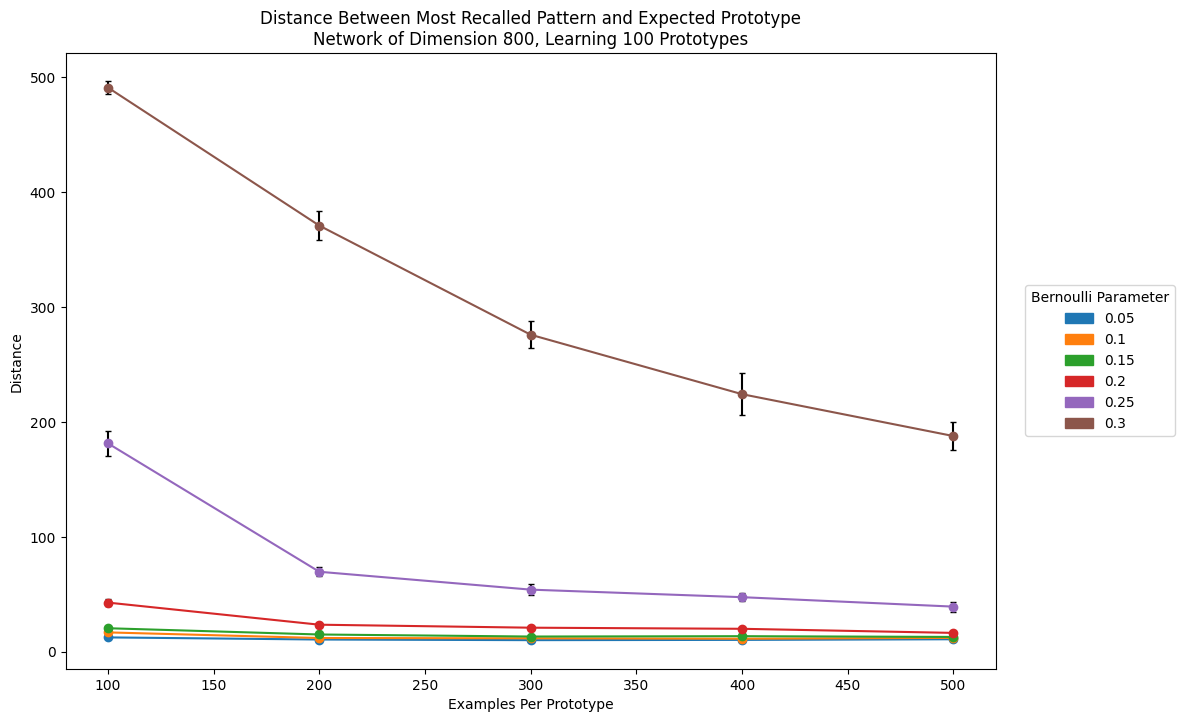}
        \caption{At Prototype Capacity, \(\alpha = 0.125\)}
        \label{Fig:DistanceAtCapacity}
    \end{subfigure}
    \caption{Distance between most recalled state and nearest representative state in different phases of the Hopfield network (below, at, and above prototype capacity).}
\end{figure}

Figure \ref{Fig:DistanceBelowCapacity} shows the average Manhattan distance between the most recalled and nearest representative state when the number of presented prototypes is well below the network's capacity. This region is most similar to the long-range spin-glass in the retrieval phase, where the global minima are learned states. Here, we see similar results as even for large Bernoulli parameters (large noise in the training examples), the attractor stabilized in the network is extremely close to the representative state (note the y-axis range). 

Figure \ref{Fig:DistanceAboveCapacity} shows the distance for a network well above the prototype capacity. As expected, the network fails to stabilize an attractor anywhere near the representative state, and increasing the number of examples does not help. Section \ref{Section:ProportionOfProbesData} also shows the attractors selected here are very weak, meaning we have only spurious states and not prototype formation. We believe the relatively small distances to the representative state are due to random spurious states appearing, in accordance to \citep{McEliece1987}, rather than any prototype formation.

Figure \ref{Fig:DistanceAtCapacity} shows the distance for a network at prototype capacity, or just below it. We see the network is capable of stabilizing an attractor close to the representative state for smaller Bernoulli parameters, although the number of examples may have to be large. For larger Bernoulli parameters (e.g. \(p\geq0.3\)) the distance to the nearest prototype is still significant; the prototype formation is disrupted but not lost. The prototype capacity appears to be influenced not only by the size of the network but also by the properties of the training data.

The above data consists of only some of our experimentation. All of our data may be found in Appendix \ref{Apnx:AllData}.

\subsection{Proportion of Probes Recalling the Most Recalled State} \label{Section:ProportionOfProbesData}

\begin{figure}[H]
    \centering
    \begin{subfigure}{\textwidth}
        \centering
        \includegraphics[width=\textwidth]{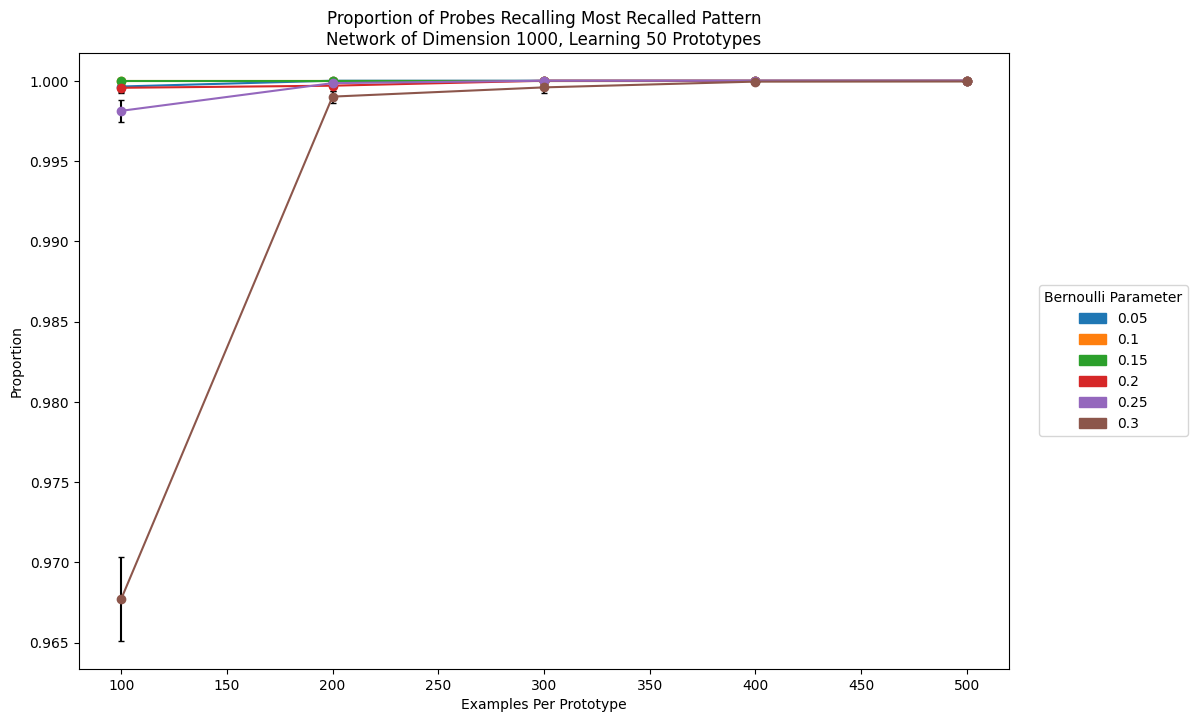}
        \caption{Below Prototype Capacity, \(\alpha = 0.05\)}
        \label{Fig:RecallBelowCapacity}
    \end{subfigure}
    \hfill
    \begin{subfigure}{\textwidth}
        \centering
        \includegraphics[width=\textwidth]{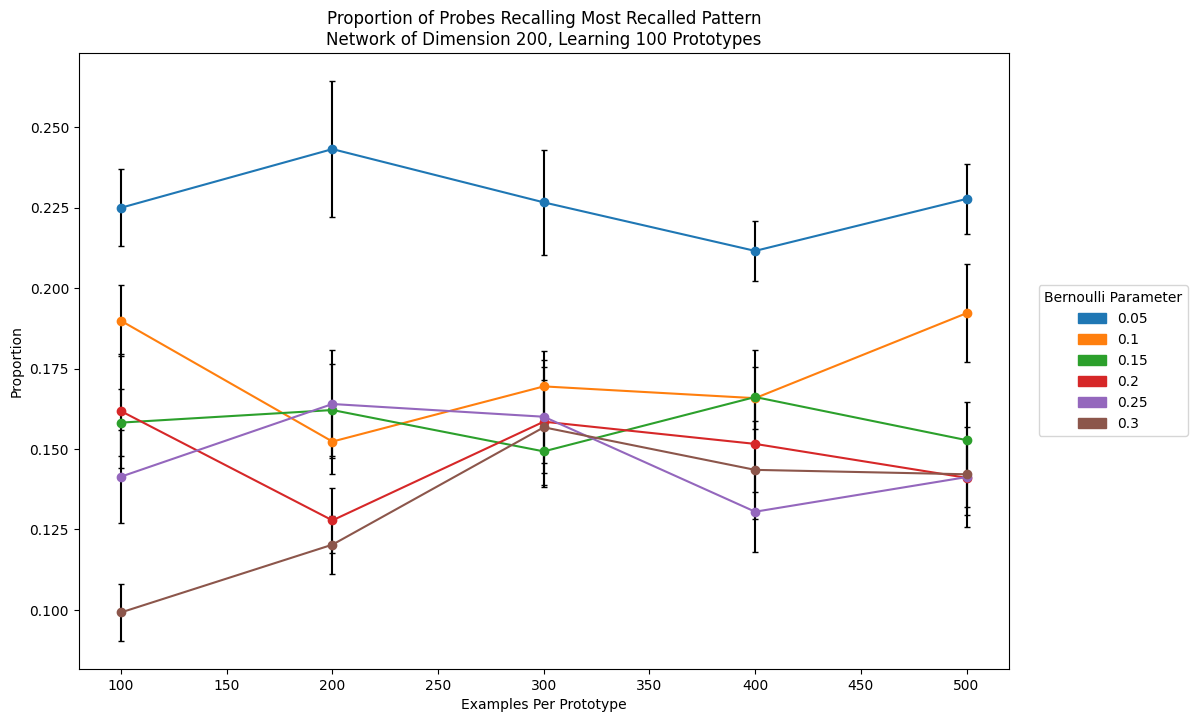}
        \caption{Above Prototype Capacity, \(\alpha = 0.5\)}
        \label{Fig:RecallAboveCapacity}
    \end{subfigure}
\end{figure}
\begin{figure}[H]\ContinuedFloat
    \centering
    \begin{subfigure}{\textwidth}
        \centering
        \includegraphics[width=\textwidth]{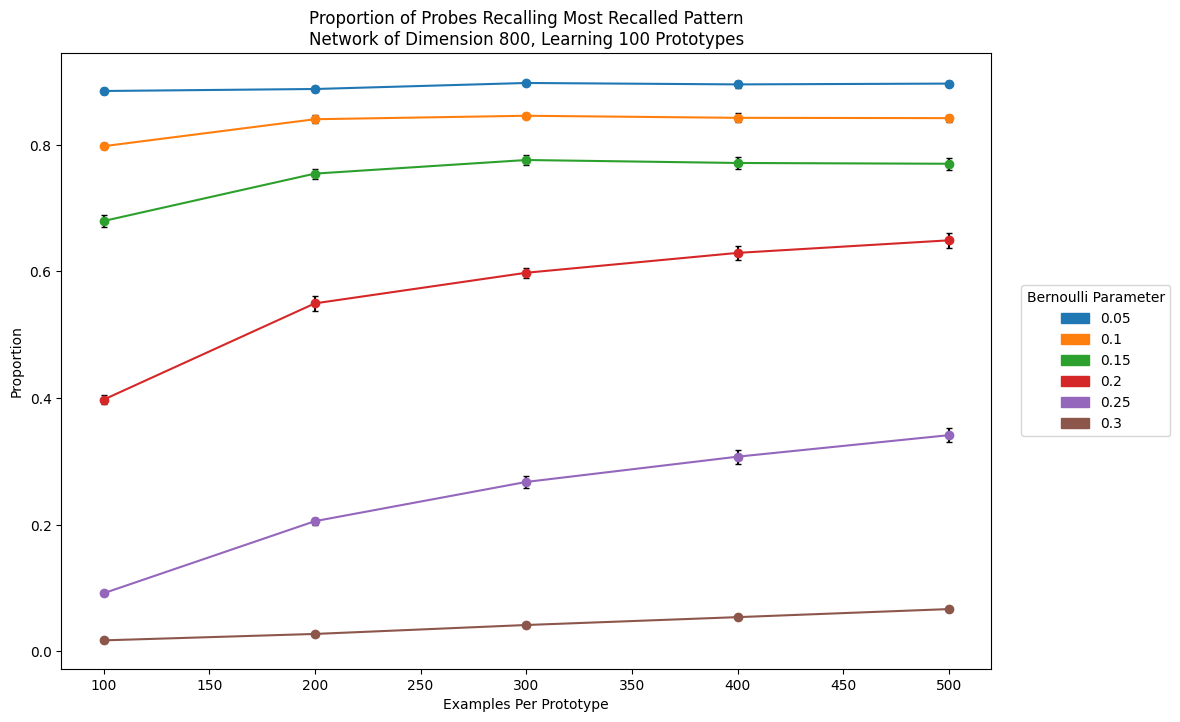}
        \caption{At Prototype Capacity, \(\alpha = 0.125\)}
        \label{Fig:RecallAtCapacity}
    \end{subfigure}
    \caption{Proportion of probes recalling the most recalled state in different phases of the Hopfield network (below, at, and above prototype capacity).}
\end{figure}

Figure \ref{Fig:RecallBelowCapacity} shows the proportion of probes that recall a most recalled state. If this value is high, closer to \(1.0\) then the formed prototype dominates the nearby attractor space. For Figure \ref{Fig:RecallBelowCapacity} this value is indeed very high, even for larger Bernoulli parameters (again, see the y-axis range). This shows that networks in the ``prototype-retrieval phase'' the formed prototype states are dominant and, from Section \ref{Section:DistanceData}, are located where we expect. We can claim with confidence that the network has formed the representative state as a prototype.

Figure \ref{Fig:RecallAboveCapacity} shows the proportions for a network storing well above the prototype capacity. In contrast to Figure \ref{Fig:RecallBelowCapacity}, the proportion of probes is much lower for all values of the Bernoulli parameter and the number of examples presented. Since the most recalled state does not dominate the nearby attractor space, and since the most recalled state does not represent a subset of as seen in Section \ref{Section:DistanceData}, we can say the network has failed to form any prototypes.

Finally, Figure \ref{Fig:RecallAtCapacity} shows the proportion for a network at capacity, or just below it. More interesting than Figure \ref{Fig:RecallBelowCapacity} and \ref{Fig:RecallAboveCapacity} we see the proportion increases with the number of presented examples, although never reaches \(1.0\). For larger Bernoulli parameters the proportion is small, sometimes smaller than in the ``above capacity'' region (yet more evidence we are observing a different phenomenon in the ``above capacity'' region). This again indicates prototype formation near the network capacity, although the accuracy and strength is somewhat influenced by the number of prototypes, the number of examples, and the Bernoulli parameter.

The above data consists of only some of our experimentation. All of our data may be found in Appendix \ref{Apnx:AllData}.

\subsection{Energy Profiles of States by Class}

\begin{figure}[H]
    \centering
    \begin{subfigure}{\textwidth}
        \centering
        \includegraphics[width=\textwidth]{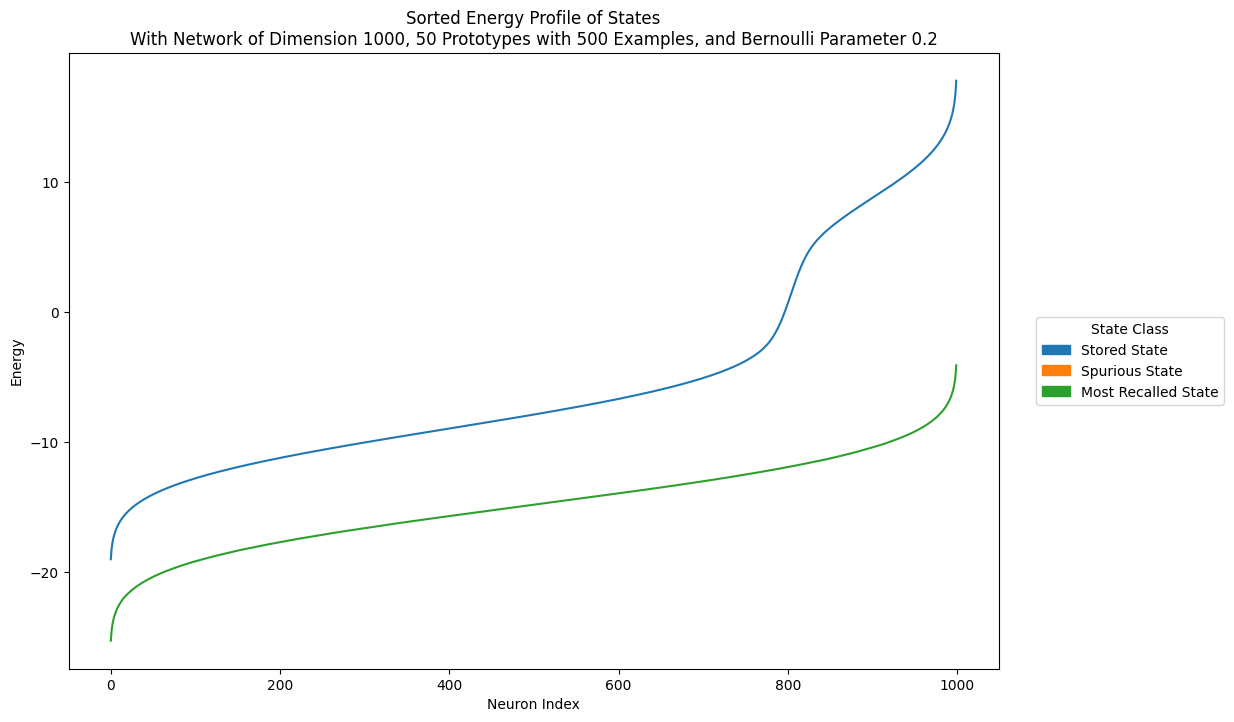}
        \caption{Below Prototype Capacity, \(\alpha = 0.05\)}
        \label{Fig:EnergyBelowCapacity}
    \end{subfigure}
    \hfill
    \begin{subfigure}{\textwidth}
        \centering
        \includegraphics[width=\textwidth]{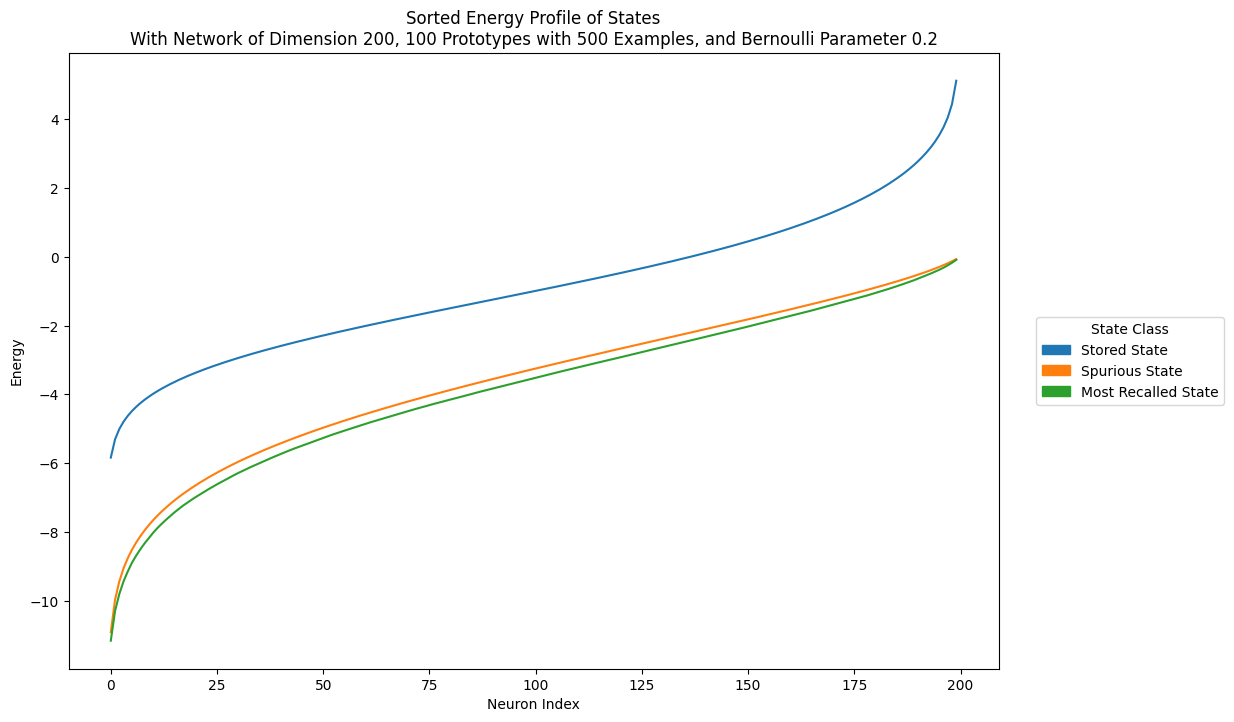}
        \caption{Above Prototype Capacity, \(\alpha = 0.5\)}
        \label{Fig:EnergyAboveCapacity}
    \end{subfigure}
\end{figure}
\begin{figure}[H]\ContinuedFloat
    \centering
    \begin{subfigure}{\textwidth}
        \centering
        \includegraphics[width=\textwidth]{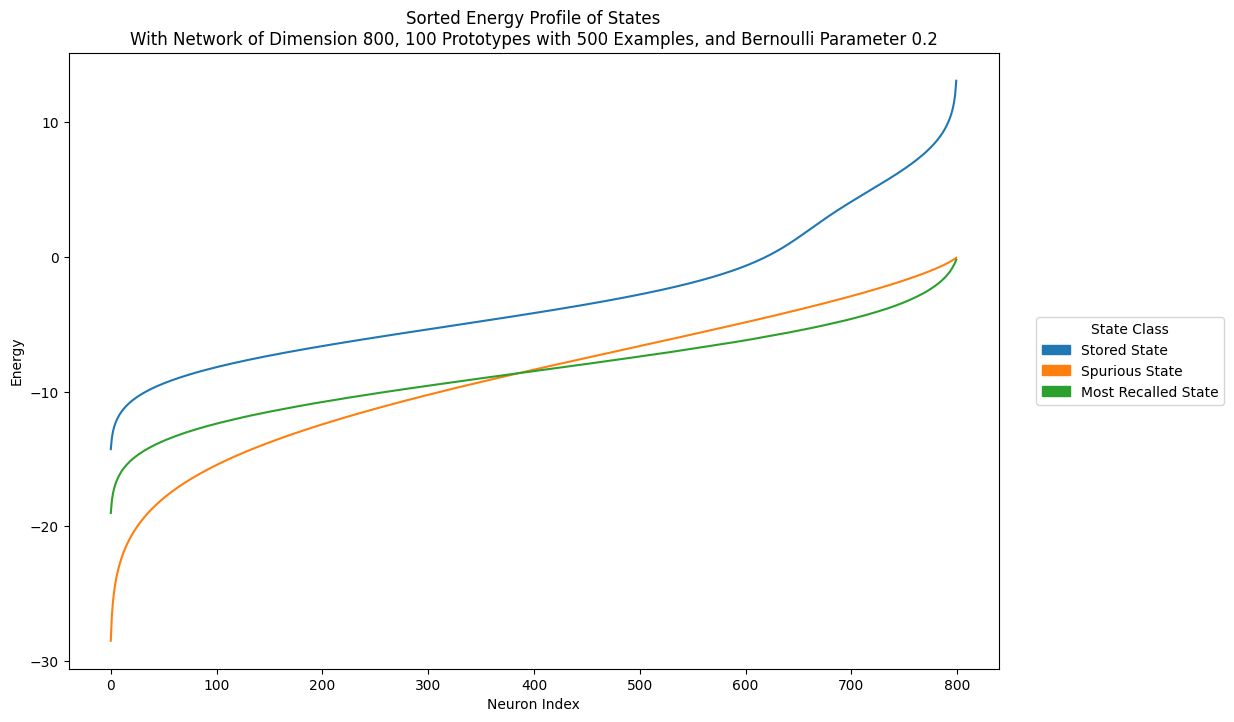}
        \caption{At Prototype Capacity, \(\alpha = 0.125\)}
        \label{Fig:EnergyAtCapacity}
    \end{subfigure}
    \caption{Sorted energy profile of different classes of states in different phases of the Hopfield network (below, at, and above prototype capacity).}
\end{figure}

In the Hopfield network, the energy of any individual neuron determines the stability of that neuron. By sorting the energies of a state we can look at a sort of fingerprint --- how many neurons are very stable (more negative energy) and how many are unstable (energy above zero). Comparing the different energy profiles between classes may be useful for determining why some states are more stable or dominant than others. Energy profiles have also been used in a similar context to distinguish learned, spurious, and prototype states \citep{Gorman2017}, demonstrating this method is not without merit. These previous works analyzed only the stable learned states, while we include energy profiles for all learned states regardless of stability. We present a slightly different class of states as we found vanishingly few stable learned states in even the least overloaded networks, and none in the more overloaded ones.

Figure \ref{Fig:EnergyBelowCapacity} shows the energy profile for a network well below the prototype capacity. We see the learned states have many neurons above zero energy, and are hence unstable. This is to be expected as we are greatly exceeding the traditional capacity of the Hopfield network, so no learned state (or very few) will stabilize. The most recalled states, which in this regime we now know are prototype formations, stabilize with a much flatter energy profile, with the most unstable neuron still having less than zero energy. This means the energy of the most stable neuron is much closer to that of the most unstable neuron, rather than the state having a few very stable neurons and many that are unstable like in the learned states. Interestingly, in this phase of the network, no spurious states were recalled, and hence their energy profiles are absent.

Figure \ref{Fig:EnergyAboveCapacity} shows the energy profiles for a network well above the prototype capacity. While there is little change to the learned states (some neurons are still unstable, leading to an unstable state as expected), we now find spurious states appearing. The energy profile of the spurious and most recalled states are extremely similar, again because no prototype formation occurred and only spurious states emerge.

Figure \ref{Fig:EnergyAtCapacity} shows the energy profile for a network at, or just below, the prototype capacity. The learned states are again unstable, but we can now distinguish the spurious and most recalled states. While the spurious states have a \textit{very} stable, low-energy ``core'' of neurons, all other neurons are much less stable. In comparison, the most recalled states have a flatter profile, showing that all neurons have about the same stability. This shows that most recalled states, when formed as prototypes, are more robust to disruption than spurious states; a small perturbation to the weight matrix may be enough to tip the most unstable neurons in a spurious state into positive energy, while it would take a relatively large perturbation to destabilize a prototype state.

\section{Conclusion}
We conduct a substantial literature review of the Hopfield network and prototype formation, including works from computer science, statistical physics, and psychology. We provide an overview of the research relating to the nature of stability of learned patterns, the capacity of the network, and the nature of spurious patterns. We also look at Modern Hopfield Networks/Dense Associative Memories and look at the new claims about prototype formation in these networks, contrasting with our work.

We give a theoretical analysis of prototype formation in the Hopfield network under Hebbian learning. We find a stability condition for prototype attractors based on the size of the subset defining the prototype, as well as the similarity of the patterns in that subset. This analysis reduces to the traditional stability condition for a single pattern in the Hopfield network under Hebbian learning, which we generalize to predict the probability of prototype formation with respect to various conditions of the training data.

We conduct experiments to empirically investigate prototype formation. We investigate the formation of a single prototype in the Hopfield network, finding results that corroborate the theory. We find the empirical results fall short of the theory in some aspects, which we believe is due to finitely many presented examples and small network dimensions, violating some assumptions of the theory.

We also conduct experiments on a more practical example, successfully forming multiple prototypes in a single network. We investigate the properties of these prototypes with respect to the number of prototypes formed, the number of examples presented, and the amount of noise in the examples. The strength of prototype formation is indirectly measured by probing the basin of attraction. We find a positive trend between prototype strength and the number of presented examples. We find negative trends between prototype strength and example noise, and prototype strength and the number of learned prototypes. Finally, we investigate the energy profile of learned, spurious, and prototype patterns in an effort to explain why prototype formation dominates the network.

\appendix

\section{Comprehensive Results} \label{Apnx:AllData}

\begin{figure}[H] 
    \begin{center}
        \includegraphics[width=\textwidth]{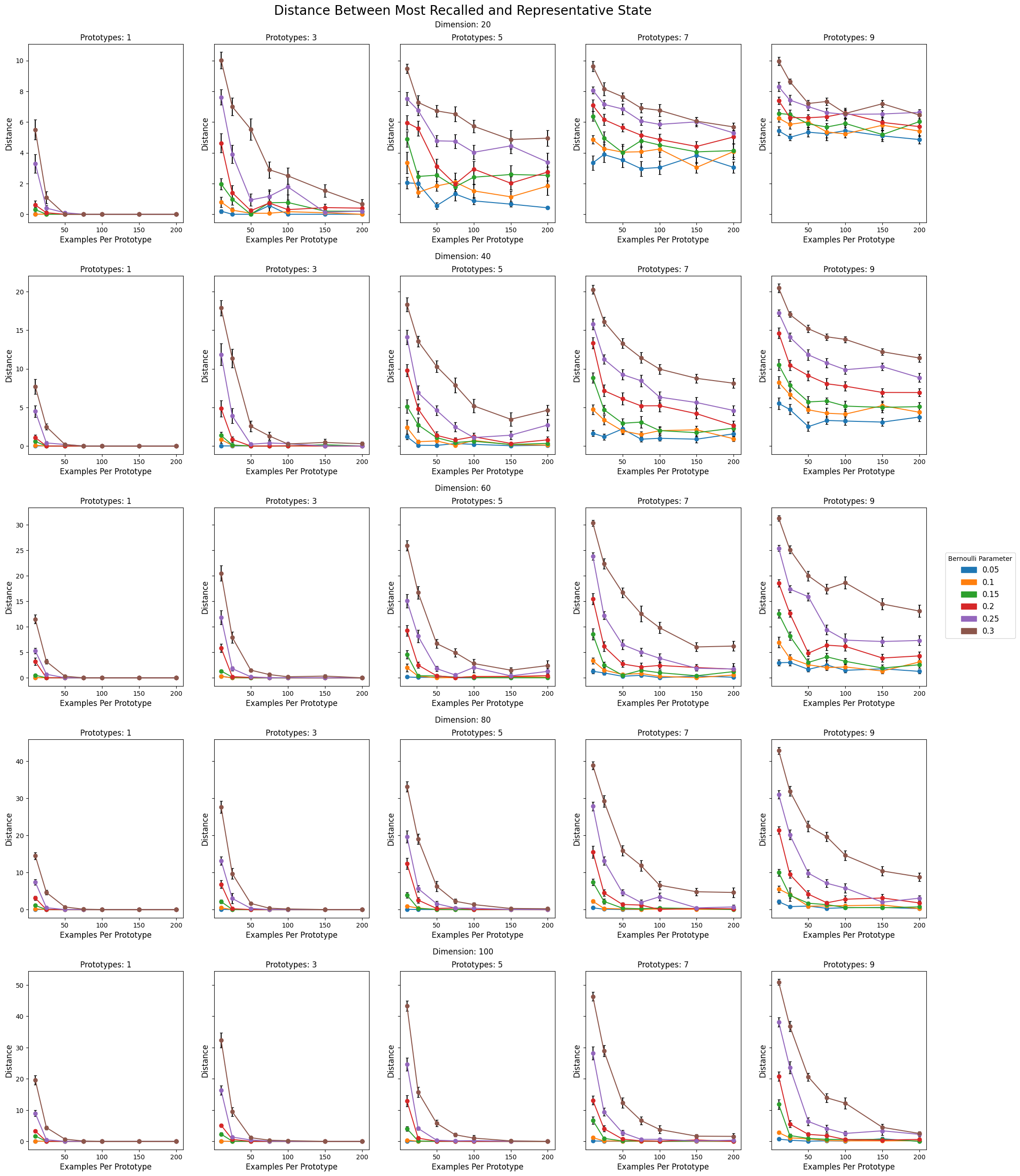}
    \end{center}
    \caption{Data from a grid search of various network dimensions (up to 100 neurons), number of prototypes, number of examples, and Bernoulli parameter, measuring the distance between the most recalled pattern and the expected prototype.}
\end{figure}

\begin{figure}[H] 
    \begin{center}
        \includegraphics[width=\textwidth]{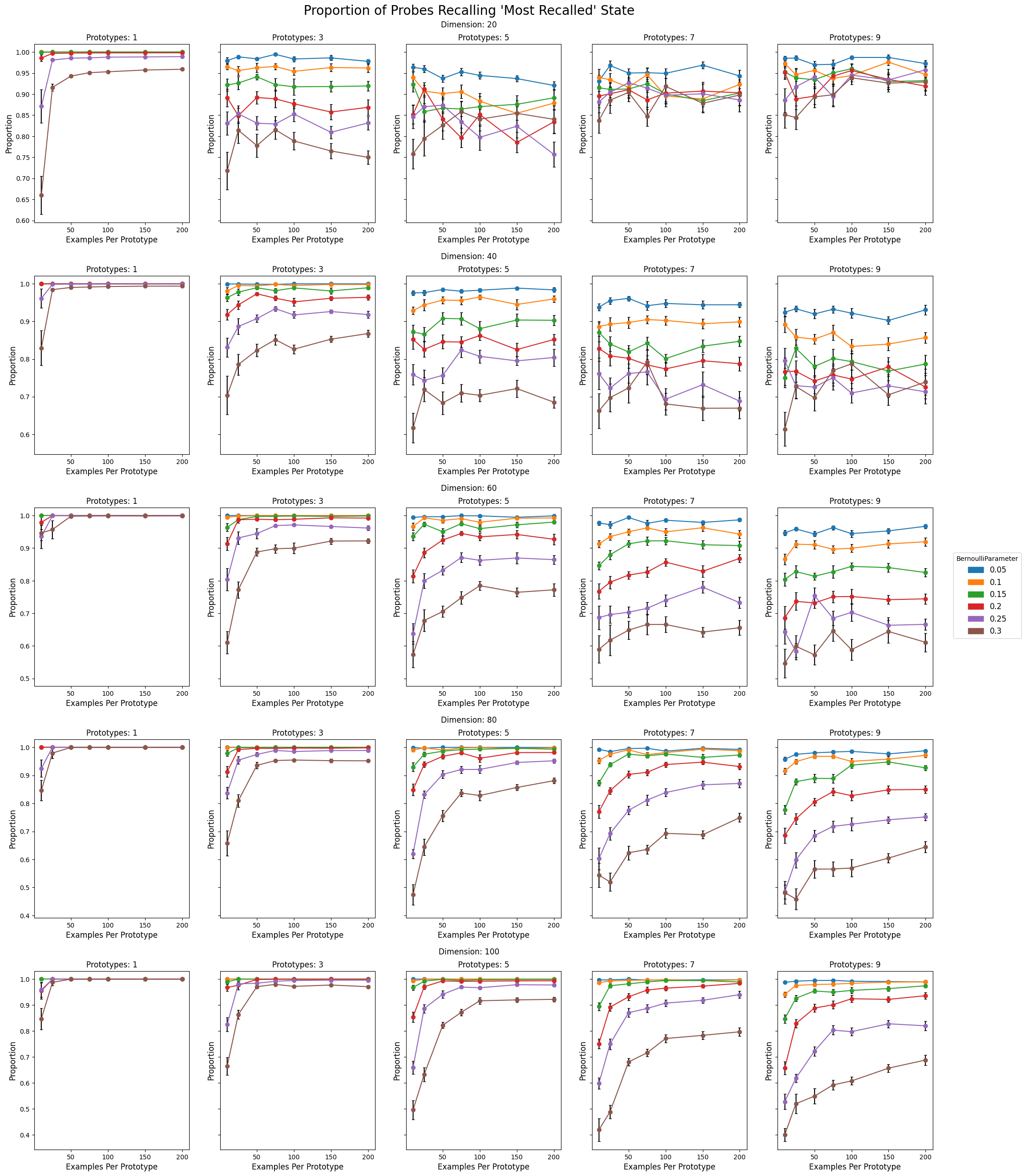}
    \end{center}
    \caption{Data from a grid search of various network dimensions (up to 100 neurons), number of prototypes, number of examples, and Bernoulli parameter, measuring the proportion of probes recalling the most recalled pattern, in effect measuring prototype dominance over the attractor space.}
\end{figure}

\begin{figure}[H] 
    \begin{center}
        \includegraphics[width=\textwidth]{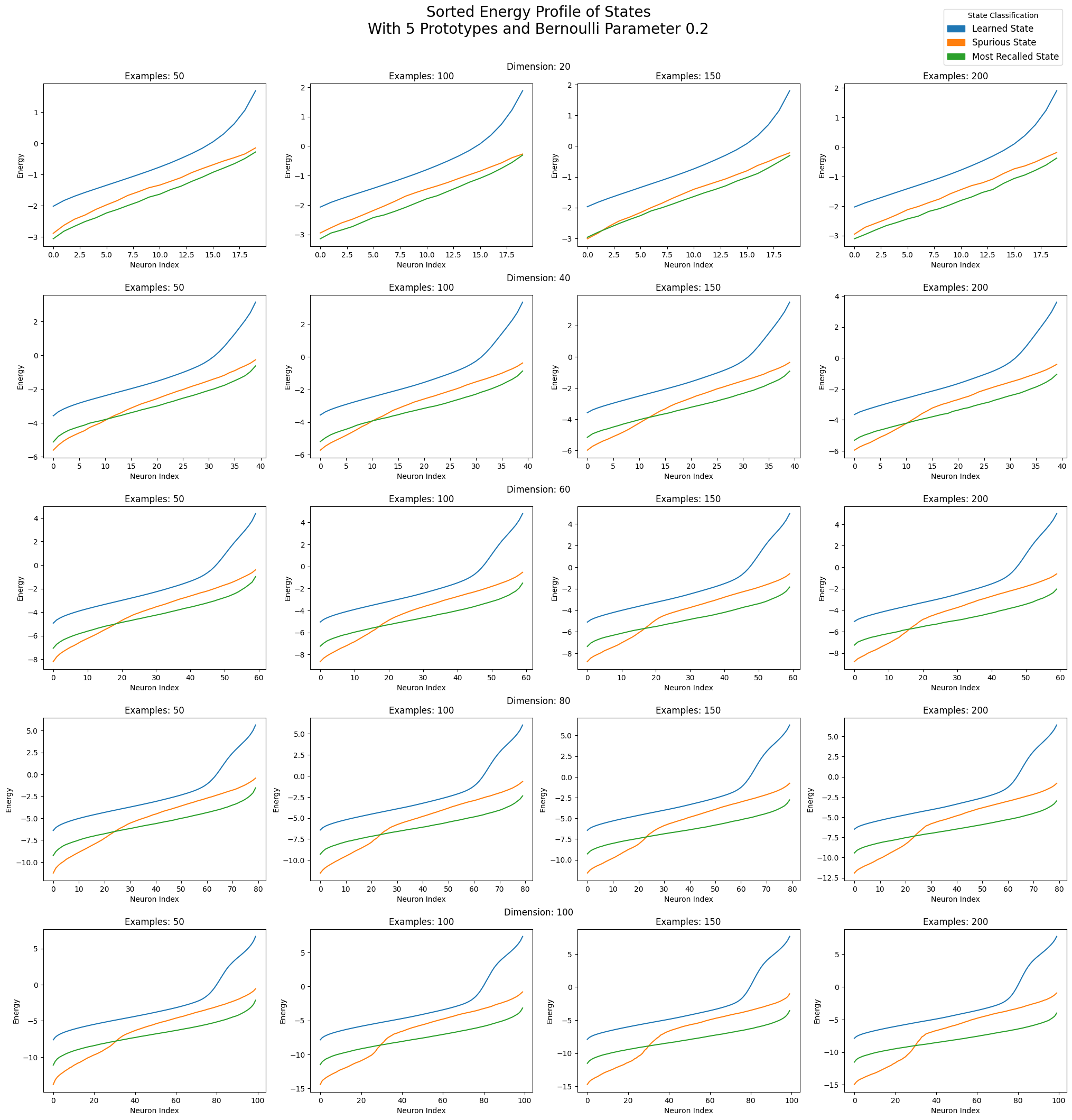}
    \end{center}
    \caption{Data from a grid search of various network dimensions (up to 100 neurons) and number of examples, with the number of prototypes and Bernoulli parameter kept fixed, measuring the energy profiles of the target, spurious, and prototype patterns.}
\end{figure}

\begin{figure}[H] 
    \begin{center}
        \includegraphics[width=\textwidth]{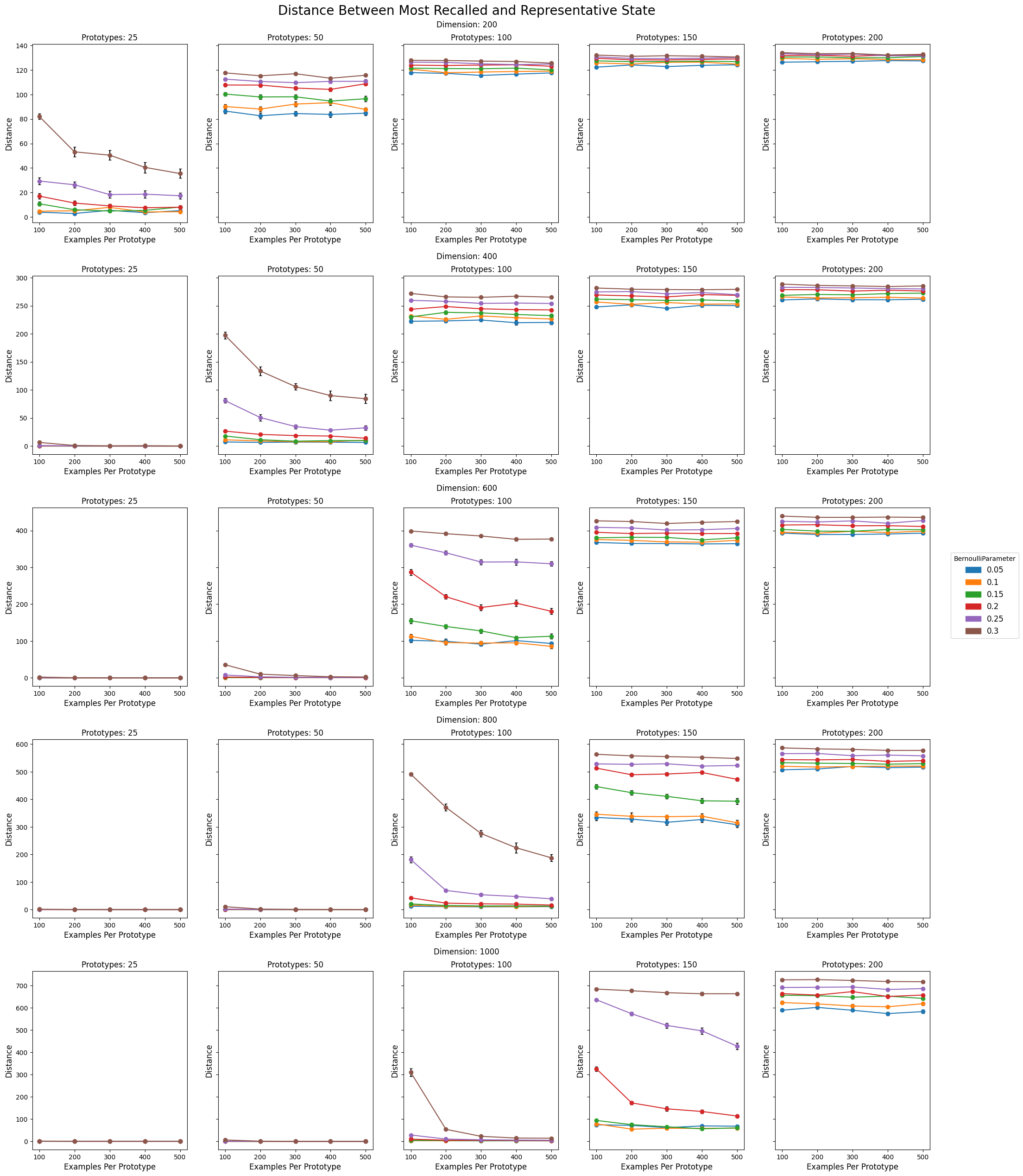}
    \end{center}
    \caption{Data from a grid search of various network dimensions (up to 1000 neurons), number of prototypes, number of examples, and Bernoulli parameter, measuring the distance between the most recalled pattern and the expected prototype.}
\end{figure}

\begin{figure}[H] 
    \begin{center}
        \includegraphics[width=\textwidth]{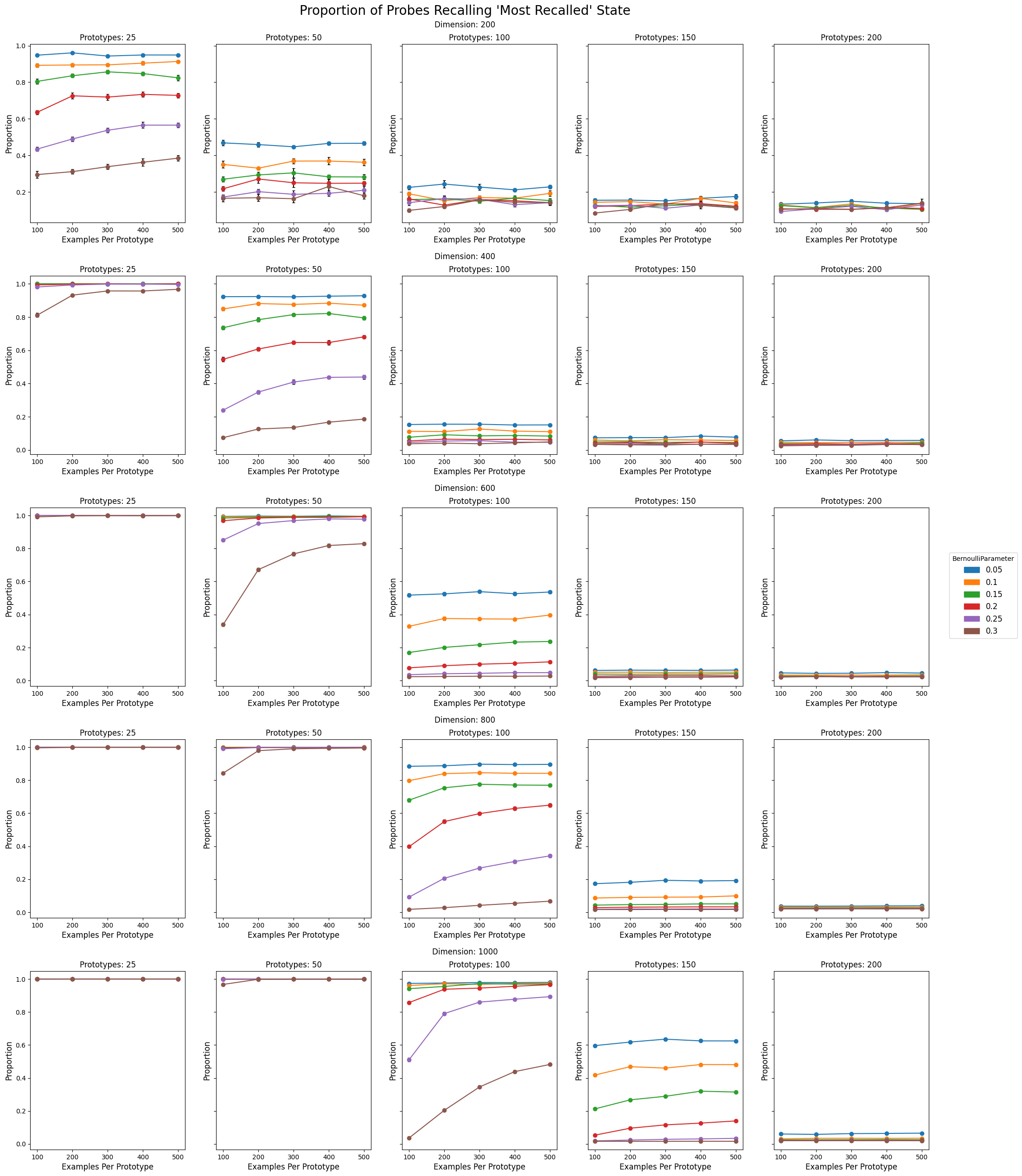}
    \end{center}
    \caption{Data from a grid search of various network dimensions (up to 1000 neurons), number of prototypes, number of examples, and Bernoulli parameter, measuring the proportion of probes recalling the most recalled pattern, in effect measuring prototype dominance over the attractor space.}
\end{figure}

\begin{figure}[H] 
    \begin{center}
        \includegraphics[width=\textwidth]{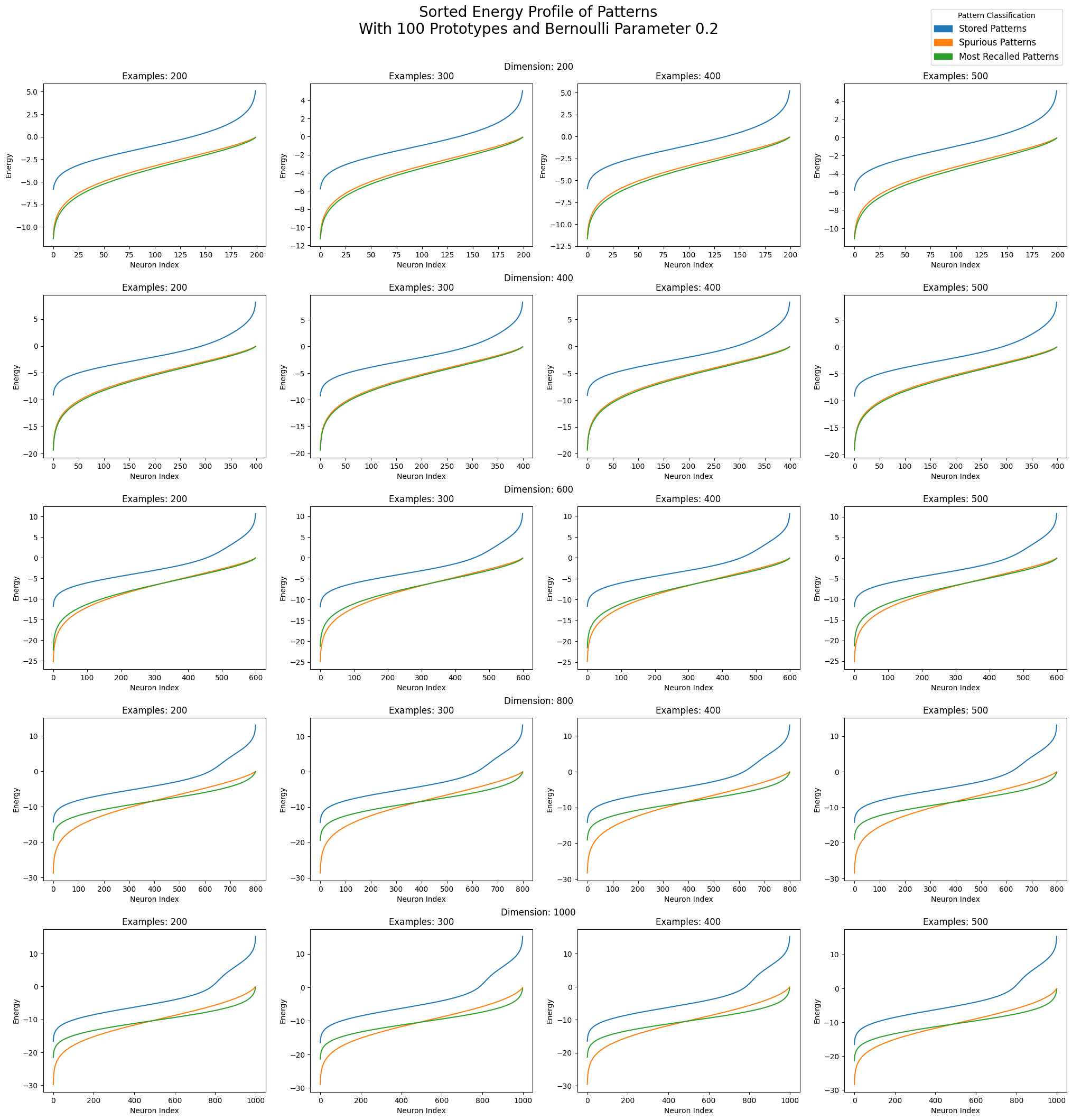}
    \end{center}
    \caption{Data from a grid search of various network dimensions (up to 1000 neurons) and number of examples, with the number of prototypes and Bernoulli parameter kept fixed, measuring the energy profiles of the target, spurious, and prototype patterns.}
    \label{Fig:AppendixEnergyGrid}
\end{figure}

\begin{figure}[H] 
    \begin{center}
        \includegraphics[width=\textwidth]{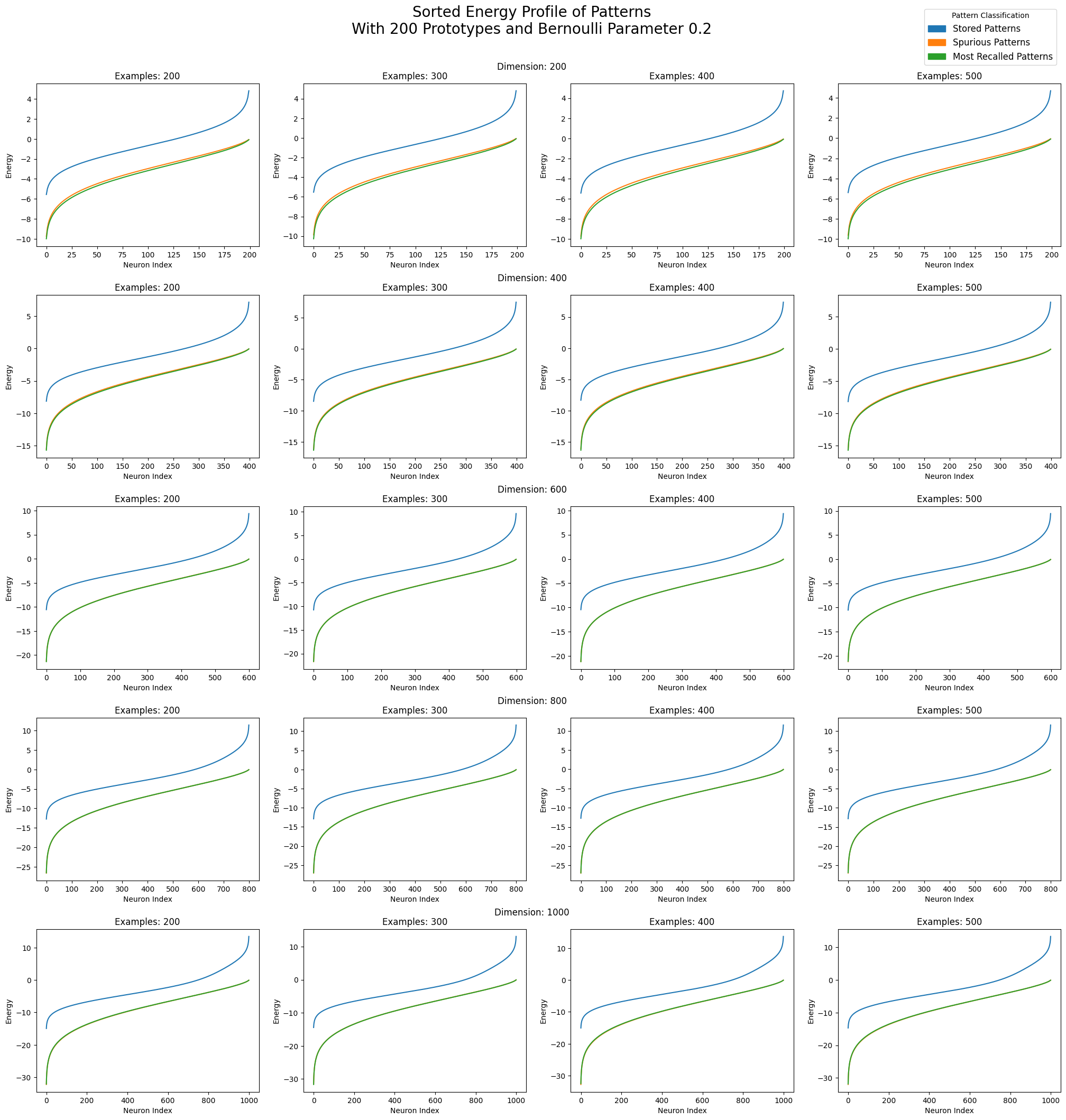}
    \end{center}
    \caption{Data from a grid search of various network dimensions (up to 1000 neurons) and number of examples, with the number of prototypes and Bernoulli parameter kept fixed, measuring the energy profiles of the target, spurious, and prototype patterns. In comparison to Figure \ref{Fig:AppendixEnergyGrid}, we increase the number of prototypes.}
\end{figure}

\begin{figure}[H] 
    \begin{center}
        \includegraphics[width=\textwidth]{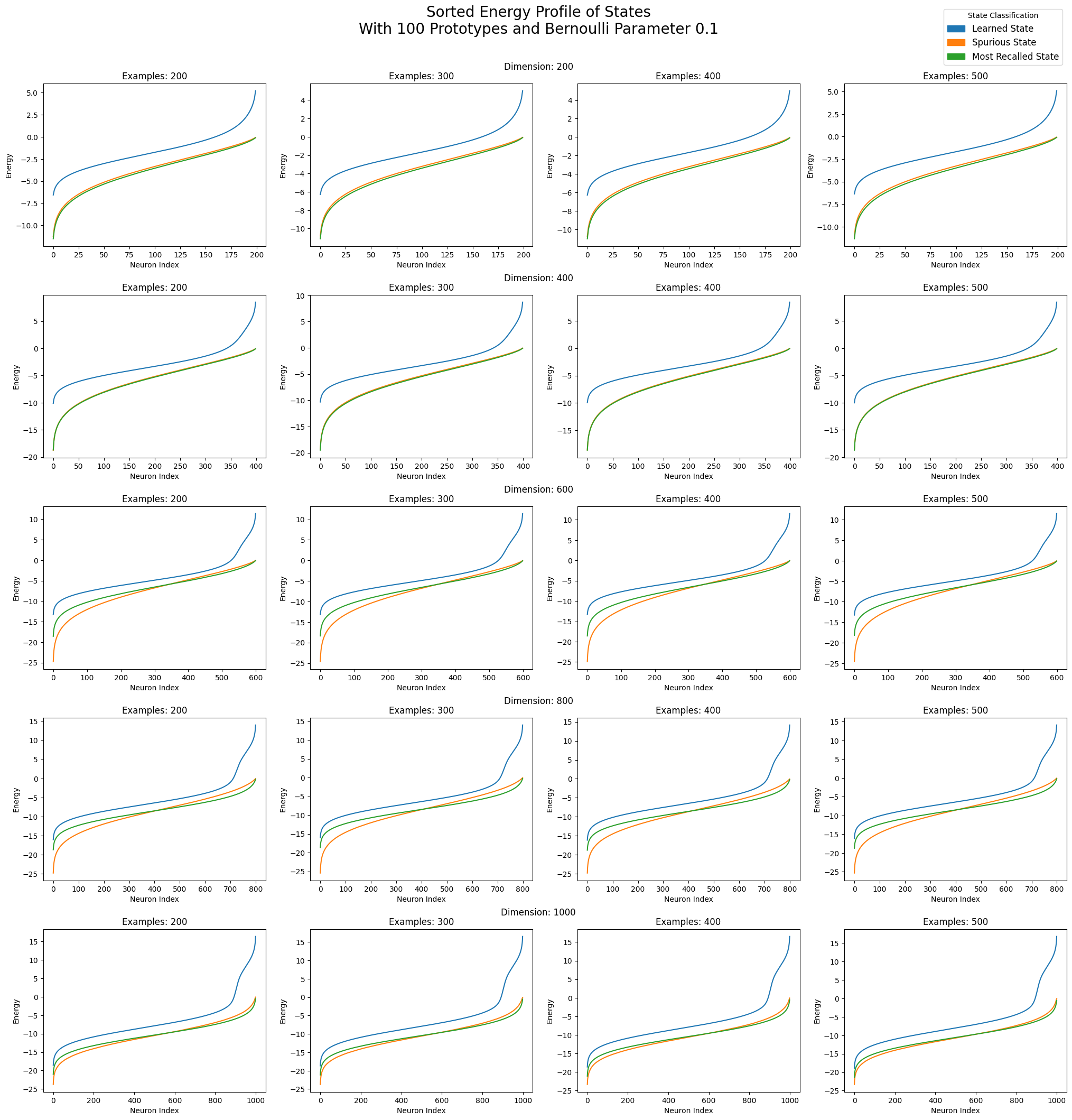}
    \end{center}
    \caption{Data from a grid search of various network dimensions (up to 1000 neurons) and number of examples, with the number of prototypes and Bernoulli parameter kept fixed, measuring the energy profiles of the target, spurious, and prototype patterns. In comparison to Figure \ref{Fig:AppendixEnergyGrid}, we decrease the Bernoulli parameter.}
\end{figure}

\bibliographystyle{apalike}
\bibliography{bibliography}

\end{document}